\pdfoutput=1
\documentclass[10pt,twocolumn,letterpaper]{article}

\usepackage{cvpr}
\usepackage{times}
\usepackage{epsfig}
\usepackage{graphicx}

\usepackage{enumitem}
\usepackage{caption}
\usepackage{dsfont}
\usepackage{wrapfig}
\usepackage{amsfonts,amsmath,amssymb,amsthm}
\usepackage{bm,nicefrac}
\DeclareMathSymbol{\shortminus}{\mathbin}{AMSa}{"39}
\usepackage[bottom]{footmisc}
\long\def\IGNORE#1{} \long\def\COMMENT#1{}
\long\def\MENTOUT#1{} \long\def\COMMENT#1{}

\usepackage[labelfont=bf, labelsep=period]{caption} 

\usepackage[pagebackref=true,breaklinks=true,letterpaper=true,colorlinks,bookmarks=false]{hyperref}

\cvprfinalcopy 

\def\cvprPaperID{2759} 

\usepackage{color}
\definecolor{darkorange}{rgb}{1.0, 0.55, 0.0}
\definecolor{pink}{RGB}{219, 48, 122}
\newcommand{\LJ}[1]{\textcolor{red}{[\textbf{LJ}:~#1]}} 
\newcommand{\JS}[1]{\textcolor{blue}{[\textbf{JS}:~#1]}}

\newcommand{\hyunsoo}[1]{\textcolor{pink}{[\textbf{hyunsoo}:~#1]}}

\begin{document}

\title{Animating Human from a Single Image in the Wild \\\hyunsoo{I am not a big fan of using ``in the wild'' unless there is a critical component in the method to handle in the wild images. I would go with a title that includes goal+method, e.g., Pose guided human animation from a single image..}\LJ{The reason why we used ``in the wild" is that we would like to emphasize that our method can generalize well to unseen scenes including unseen background and unseen person appearance}}
\title{Pose-Guided Human Animation from a Single Image in the Wild}

\author{
Jae Shin Yoon$^\dagger$
\hspace{5mm}Lingjie Liu$^\sharp$
\hspace{5mm}Vladislav Golyanik$^\sharp$
\hspace{5mm}Kripasindhu Sarkar$^\sharp$
\vspace{1mm}
\\
\hspace{0mm}Hyun Soo Park$^\dagger$
\hspace{10mm}Christian Theobalt$^\sharp$
\vspace{3mm}
\\
\hspace{-0mm}$^\dagger$University of Minnesota
\hspace{10mm}
$^\sharp$Max Planck Institute for Informatics, SIC \\
}

\twocolumn[{
\maketitle
	\begin{center}
		\centering
		\vspace{-6mm}
	 \includegraphics[width=1\linewidth]{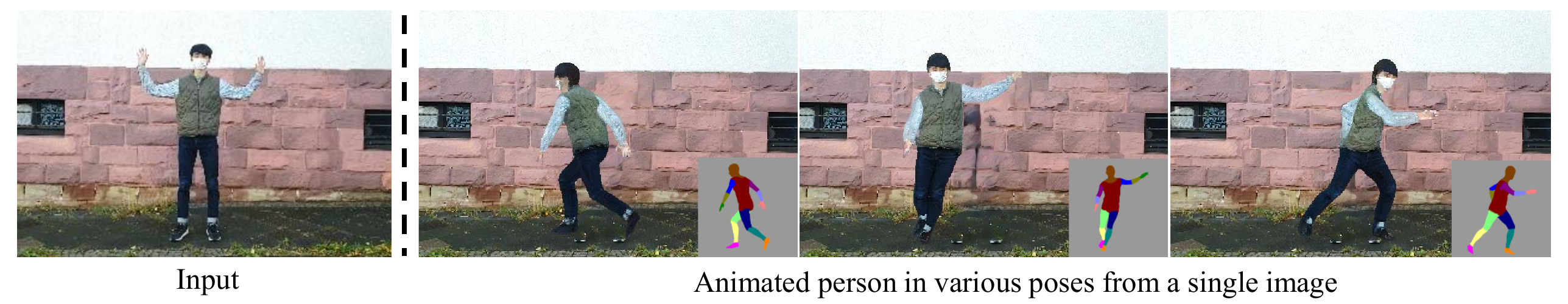}
	 \vspace{-7mm}
	\captionof{figure}{\small We present a new method to synthesize a sequence of animated human images from a single image. The synthesized images are controlled by the poses as shown in the inset image. It is capable of generating full appearance of the person at diverse poses, reflecting the input foreground and background in the presence of occlusion and 3D shape deformation, \textit{e.g.,} the occluded texture of the back. 
	} 
\end{center}	
	\label{fig:teaser}
}]

\begin{abstract}
\vspace{-3mm}

We present a new pose transfer method for synthesizing a human animation from a single image of a person controlled by a sequence of body poses. Existing pose transfer methods exhibit significant visual artifacts when applying to a novel scene, resulting in temporal inconsistency and failures in preserving the identity and textures of the person. To address these limitations, we design a compositional neural network that predicts the silhouette, garment labels, and textures. Each modular network is explicitly dedicated to a subtask that can be learned from the synthetic data. At the inference time, we utilize the trained network to produce a unified representation of appearance and its labels in UV coordinates, which remains constant across poses. The unified representation provides an incomplete yet strong guidance to generating the appearance in response to the pose change. We use the trained network to complete the appearance and render it with the background. With these strategies, we are able to synthesize human animations that can preserve the identity and appearance of the person in a temporally coherent way without any fine-tuning of the network on the testing scene. Experiments show that our method outperforms the state-of-the-arts in terms of synthesis quality, temporal coherence, and generalization ability.

\vspace{-3mm}

\end{abstract}



\section{Introduction}
Being able to animate a human in everyday apparel with an arbitrary pose sequence from just a single still image opens the door to many creative applications. For example, animated photographs can be much more memorable than static images. Furthermore, such techniques not only simplify and democratize computer animation for non-experts, they can also expedite pre-visualization and content creation for more professional animators who may use single image animations as basis for further refinement. 

Tackling this problem using classical computer graphics techniques is highly complex and time consuming. A high-quality 3D textured human model needs to be reconstructed from a single image and then sophisticated rigging methods are required to obtain an animatable character. An alternative is to apply 2D character animation methods~\cite{igarashi2005rigid,jacobson2011bounded} to animate the person in the image. However, this approach cannot visualize the occluded parts of the character.

In this paper, we approach this problem using a pose transfer algorithm that synthesizes the appearance of a person at arbitrary pose by transforming the appearance from an input image without requiring a 3D animatable textured human model. Existing works on pose transfer have demonstrated promising results only when training and testing take place on the same dataset (\textit{e.g.,}~DeepFashion dataset ~\cite{liu2016deepfashion}), and some require even more restrictive conditions that testing is performed on the same person in the same environment as training.~\cite{chan2019everybody,liu2020neural,liu2019neural}. However, the domain difference between training and testing data in real applications introduces substantial quality degradation. 

\begin{figure}
	\begin{center}
    \includegraphics[width=1\linewidth]{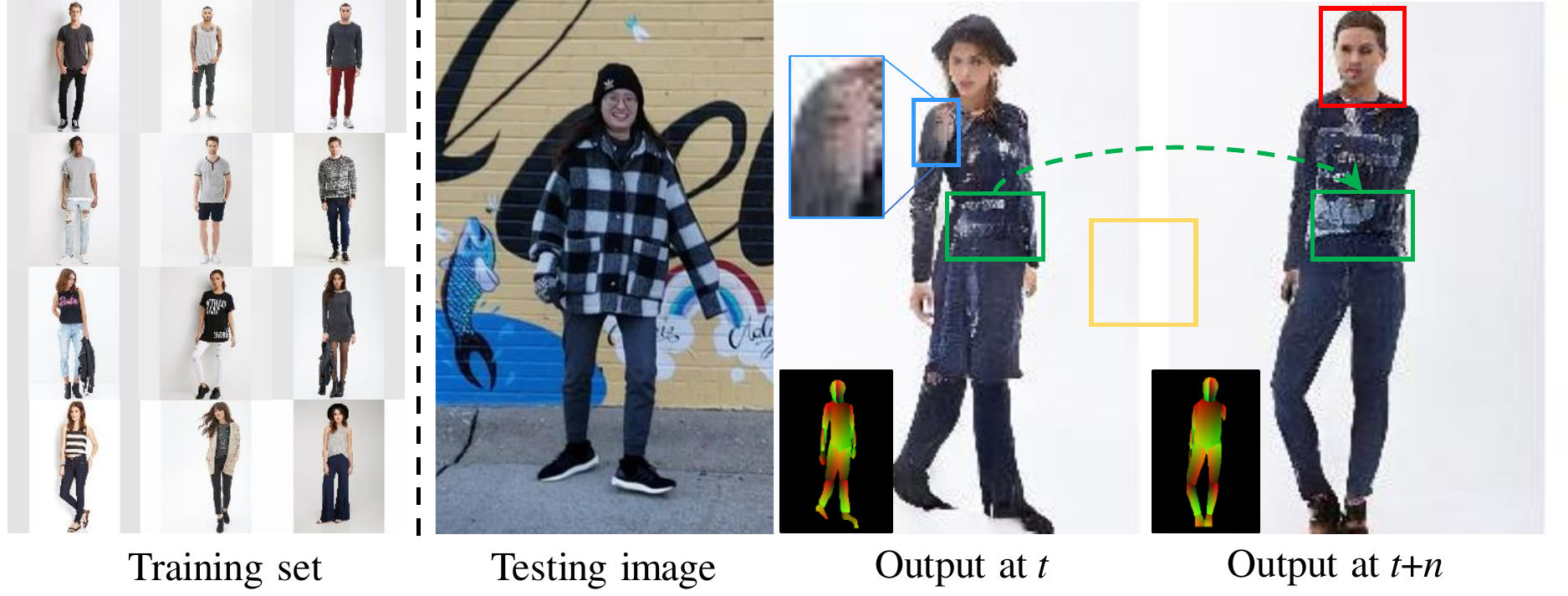}
	\end{center}	
	\vspace{-.4cm}
	\caption{\small The pose transfer results synthesized by a state-of-the-art method~\cite{sarkarneural} on an unconstrained real-world scene, where the network is trained on the Deep Fashion dataset~\cite{liu2016deepfashion}. The target body pose is shown in the inset (black). Each box represents the type of the observed artifacts such as loss of identity (red), misclassified body parts (blue), background mismatch (yellow), and temporal incoherence (green).}
	\vspace{-.4cm}
	\label{fig:motivation1}
\end{figure}

A core challenge of pose transfer lies in lack of data that span diverse poses, shapes, appearance, viewpoints, and background. This leads to limited generalizability to a testing scene, resulting in noticeable visual artifacts as shown in Fig.~\ref{fig:motivation1}. We address this challenge by decomposing the pose transfer task into modular subtasks predicting silhouette, garment labels, and textures where each task can be learned from a large amount of synthetic data. 
This modularized design makes training tractable and significantly improves result quality. Explicit silhouette prediction further facilitates animation blending with arbitrary static scene backgrounds. 

In inference phase, given the trained network from the synthetic data, we introduce an efficient strategy for synthesizing temporally coherent human animations controlled by a sequence of body poses. We first produce a unified representation of appearance and its labels in UV coordinates, which remains constant across different poses. This unified representation provides an incomplete yet strong guidance to generating the appearance in response to the pose change. We use the trained network to complete the appearance and render it with the background. Experiments show that our method significantly outperforms the state-of-the-art methods in terms of synthesis quality, temporal consistency, and generalization ability.

Our technical \textbf{contributions} include (1) a novel approach that can generate a  realistic animation of a person from a single image controlled by a sequence of poses, which shows higher visual quality and temporal coherence, and generalizes better to new subjects and backgrounds;
(2) a new compoositional pose transfer framework that produces the silhouette mask, garment labels, and textures, which makes the pose transfer tractable; (3) an effective inference method by generating a unified representation of appearance and its labels for image synthesis, enforcing temporal consistency and preserving identity in the presence of occlusion and shape deformation. 
\begin{figure*}
	\begin{center}
    \includegraphics[width=1\linewidth]{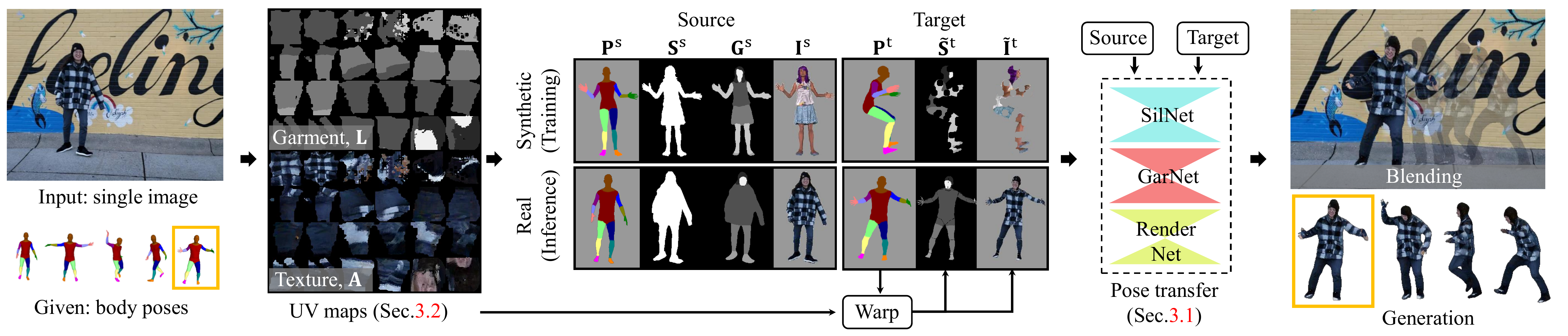}
	\end{center}	
	\vspace{-.6cm}       
	\caption{\small Overview of our approach. Given an image of a person and a sequence of body poses, we aim for generating video-realistic human animation. 
	To this end, we train a compositional pose transfer network that predicts silhouette, garment labels, and textures with synthetic data (Sec.~\ref{pose_transfer}). In inference phase, we first produce a unified representation of appearance and garment labels in the UV maps, which remains constant across different poses, and these UV maps are conditioned on our pose transfer network to generate person images in a temporally consistent way (Sec.~\ref{human_ani}). The generated images are composited with the inpainted background to produce the animation. 
	}
	\vspace{-3mm}
	\label{fig:overall}
\end{figure*}

\section{Related Work}
We review the literature for human pose transfer and its  application to the pose-guided video generation. 

\noindent\textbf{Human Pose Transfer}
Pose transfer refers to the problem of synthesizing human images with a novel user-defined pose. 
%
The conditioning pose is often captured by 2D keypoints \cite{ma2017pose,yang2020towards2,zhu2019progressive,esser2018variational,pumarola2018unsupervised,tang2019cycle} or parametric mesh \cite{liu2019liquid,ma2020unselfie,neverova2018dense,sarkarneural}.
Many recent works also use Densepose~\cite{alp2018densepose} which is the projection of SMPL model with UV parameterization in the image coordinates, as conditioning input. This enables direct warping of pixels of the input image to the spatial locations at the output with target pose ~\cite{ma2020unselfie,neverova2018dense,sarkarneural}. While the aforementioned methods produce photo-realistic results within the same dataset, they often exhibit serious artifacts on in-the-wild scenes, such as pixel blending around the boundaries between the different garment types.

To address these limitations, some recent methods use garment segmentation map, \textit{i.e.,} a label image where each pixel belongs to a semantic category such as clothing, face, and arm, as input to a neural network  ~\cite{men2020controllable,song2019unsupervised,dong2019towards,neuberger2020image}. \cite{chen2019unpaired} condition garment type, whereas \cite{balakrishnan2018synthesizing} handles each garment parts in different transformation layers to preserve the clothing style in the generated image. However, these works still do not generalize to new appearances and unseen scenes. 




Some new methods explicitly handle appearance in the occluded areas by matching their style to the visible regions.
\cite{albahar2019guided} transforms the features of the input image to a target body pose with bidirectional implicit affine transformation. \cite{han2019clothflow,ren2020deep} learn pixel-wise appearance flow in an unsupervised way based on the photometric consistency. \cite{han2019clothflow} establishes direct supervision by fitting a body model to the images. However, the predicted warping fields is often unstructured, resulting in artifacts such as shape distortion. 





\noindent\textbf{Pose-Guided Video Generation}
Since the methods for pose transfer are designed to output a single image, their application to a sequence of poses to perform pose guided video generation can exhibit temporal inconsistency. To mitigate this problem, many methods enforce explicit temporal constraints in their algorithm. 
\cite{chan2019everybody} predicts the person image in two consecutive frames. \cite{yang2018pose} conditions the temporally coherent semantics on a generative adverserial network. 
Recent video generation approaches have leveraged the optical flow prediction~\cite{wang2018vid2vid}, local affine transformation~\cite{siarohin2019first}, grid-based warping field~\cite{siarohin2019first}, body parts transformation~\cite{Zhou_2019_ICCV}, and future frame prediction~\cite{byeon2018contextvp,wang2019point} to enforce the temporal smoothness. 
\cite{liu2020neural} learns to predict a dynamic texture map that allows rendering physical effects, \textit{e.g.,} pose-dependent clothing deformation, to enhance the visual realism on the generated person.
Unfortunately, the above methods are either person-specific or requiring the fine-tuning on unseen subjects for the best performance. 
While few-shot video generation~\cite{wang2019few} addressed this generalization problem, it still requires fine-tuning on the testing scene to achieve full performance. In contrast, our method works with a single conditioning image in the wild and performs pose guided video synthesis without any fine-tuning.


\noindent\textbf{Other Related Techniques} In contrast to the data-driven neural rendering methods, few works reconstruct a personalized animatable 3D model from a single image.
For example,~\cite{alldieck2019tex2shape} reconstructs 3D geometry by regressing shape in UV-space. \cite{huang2020arch} learns an implicit function from a neural network to predict person's surface and appearance on top of a parametric body model. \cite{weng2019photo} leverages the graphics knowledge, \textit{e.g.,} skinning and rigging, to enable the character animation from a single image. 

\section{Methodology}

Our goal is to synthesize human animations from a single image guided with a sequence of arbitrary body poses. 
The overview of our pipeline is outlined in Fig.~\ref{fig:overall}. 
In the training stage, our pose transfer network learns to generate a person's appearance in different poses using a synthetic dataset which provides full ground truth. 
At inference time, given a single image of a person and a different body pose, the learned pose transfer network generates the person's appearance that is conditioned on the partial garment and texture warped from the coherent UV maps (scene-specific priors). The generated foreground is blended with the inpainted background.
In Sec.~\ref{pose_transfer}, we introduce our compositional pose transfer network, and in inference time, we use this network to create coherent UV maps and human animation from a single image in Sec.~\ref{human_ani}.

\subsection{Compositional Pose Transfer}\label{pose_transfer}
The problem of pose transfer takes as input a source image $\mathbf{I}^{\rm s}$ and a target pose $\mathbf{P}^{\rm t}$ and generates an image of the person in the target pose $\mathbf{I}^{\rm t}$: 
\begin{equation}
    \mathbf{I}^{\rm t} = f(\mathbf{P}^{\rm t}, \mathbf{I}^{\rm s}). \label{eq:direct_learning}
\end{equation}
where the superscript $\rm s$ denotes the source as the domain of the observation from the input image, and $\rm t$ denotes the target as of the generation from a body pose.

Albeit possible, directly learning the function in Eq.~(\ref{eq:direct_learning}) is challenging as requiring large amount of multiview data~\cite{ma2017pose,men2020controllable,sarkarneural}, \textit{i.e.,} it requires to learn the deformation of the shape and appearance with respect to every possible 3D pose, view, and clothing style. This results in a synthesis of unrealistic human images that are not reflective of the input testing image as shown in Fig.~\ref{fig:motivation1}. We address this challenge by leveraging synthetic data that allows us to decompose the the function into the modular functions that are responsible to predict silhouette, garment labels, and appearance, respectively. This makes the learning task tractable and adaptable to the input testing image.

\subsubsection{Dataset and Notation}
For training, we use 3D people synthetic dataset~\cite{pumarola20193dpeople} which contains 80 subjects in diverse clothing styles with 70 actions per subject captured from four different virtual views, where each action is a sequence of 3D poses. For each subject we randomly pick two instances as the source and target with different views and 3D poses. Each instance contains the following associated information:
\vspace{-2mm}
\begin{itemize}[leftmargin=*]
    \item[\tiny$\bullet$] Image: $\mathbf{I}\in\mathbb{R}^{W\times H \times 3}$ is the person image where the foreground is masked using $\mathbf{S}$.\vspace{-2mm}
    \item[\tiny$\bullet$] Pose map: $\mathbf{P}\in\{0,\cdots,14\}^{W\times H}$ is a map of body-part labels of the undressed body (14 body parts and background).\vspace{-2mm}
    \item[\tiny$\bullet$] Silhouette mask: $\mathbf{S}\in\{0,1\}^{W\times H}$ is a binary map indicating one if it belongs to the person foreground, and zero otherwise.\vspace{-2mm}
    \item[\tiny$\bullet$] Garment labels: $\mathbf{G}\in\{0,\cdots,6\}^{W\times H}$ is a map of garment labels of dressed human body, indicating hair, face, skin, shoes, garment top and bottom, and background.\vspace{-2mm}
\end{itemize}
In inference time, given $\mathbf{I}^{\rm s}$ and $\mathbf{P}^{\rm s}$, we estimate the $\mathbf{P}^{\rm s}, \mathbf{S}^{\rm s}$ and $\mathbf{G}^{\rm s}$ from $\mathbf{I}^{\rm s}$ using off-the-shelf methods, and our pose transfer network predicts $\mathbf{S}^{\rm t}, \mathbf{G}^{\rm t}$, and $\mathbf{I}^{\rm t}$. 

\begin{figure}
	\begin{center}
	\vspace{1mm}
    \includegraphics[width=0.85\linewidth]{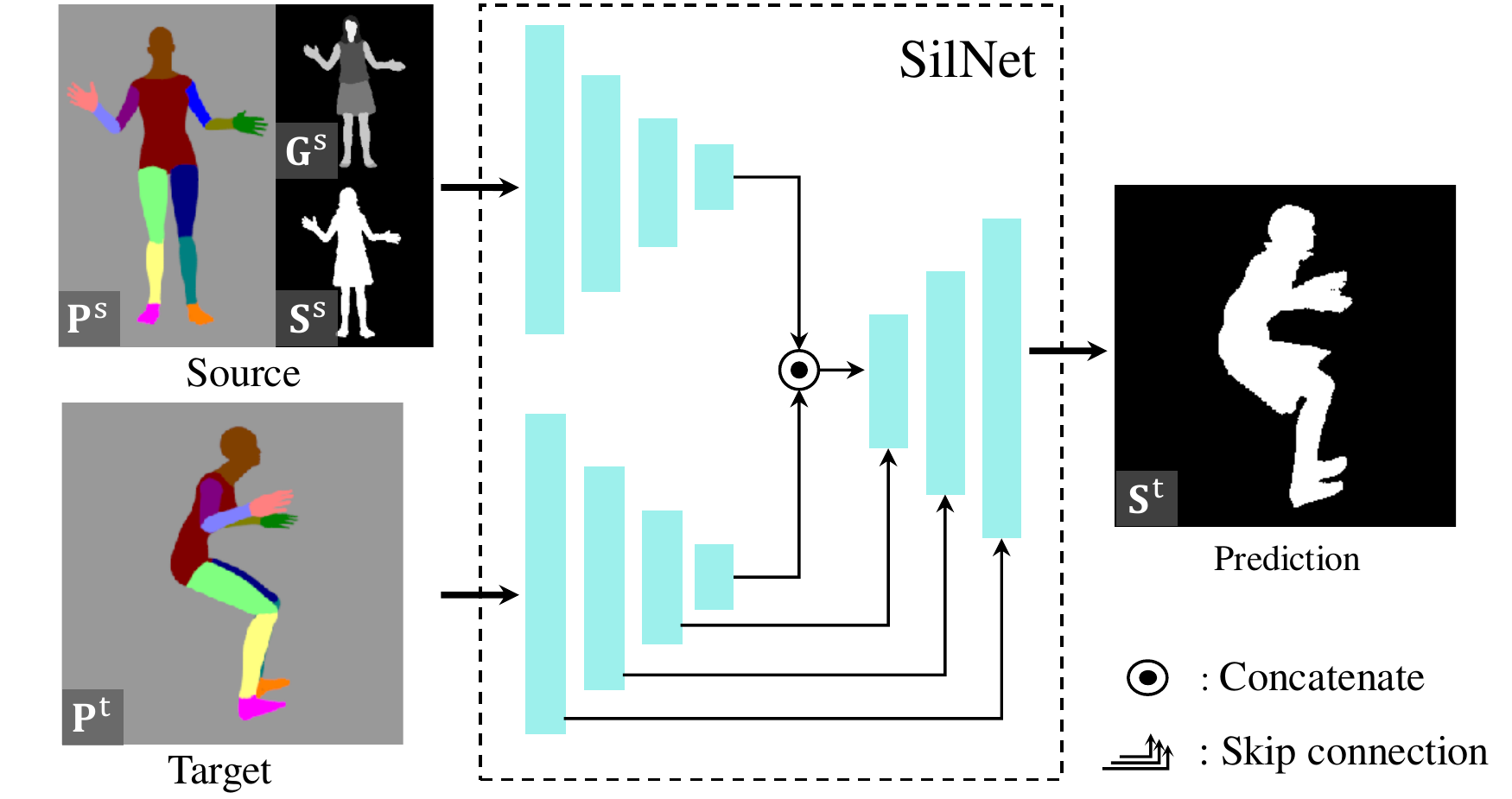}
	\end{center}	
	\vspace{-.5cm}
	\caption{\small \emph{SilNet} predicts the silhouette mask in the target pose.}
	\label{fig:MNet}
	\vspace{-2mm}
\end{figure}

\subsubsection{Silhouette Prediction} 
We predict the silhouette of the person in the target pose given the input source triplet: source pose map, silhouette, and garment label. It is designed to learn the shape deformation as a function of the pose change:
\begin{align}
\mathbf{S}^{\rm t}=f^{\rm Sil}(\mathbf{P}^{\rm t}|\{\mathbf{P}^{\rm s},\mathbf{S}^{\rm s},\mathbf{G}^{\rm s}\}).\label{mnet}
\end{align}
We use a neural network called \emph{SilNet} to learn this function. 
It has two encoders and one decoder, as shown in Fig.~\ref{fig:MNet}. One encoder encodes the spatial relationship of the body and silhouette from the source triplet, which is used to condition the silhouette generation of the target pose by mixing their latent codes. The garment labels $\mathbf{G}^{\rm s}$ provides an additional spatial cue to control the deformation, \textit{i.e.,} pixels that do not belong to garment (\textit{i.e.,} skin) less likely undergo large deformation. The features extracted from the target pose at each level are passed to the counterpart of the decoder through skip connections. We train \emph{SilNet} by minimizing the $L1$ distance of the predicted silhouette mask $\mathbf{S}^{\rm t}$ and the ground truth $\mathbf{S}_{\rm gt}^{\rm t}$:
\begin{align}
L_{\rm Sil}=\left\|\mathbf{S}^{\rm t}-\mathbf{S}_{\rm gt}^{\rm t}\right\|_1.
\end{align}
Note that, as $f^{\rm Sil}$ does not take as input the source image $\mathbf{I}^{\rm s}$, using synthetic data does not introduce the domain gap. 


\subsubsection{Garment Label Prediction}\label{garnet}
Given the source triplet and the predicted target silhouette, we predict the target garment labels $\mathbf{G}^{\rm t}$ that guide the generation of the target appearance. We take two steps. 


First, we warp the source garment labels to produce the pseudo target garment labels, $\widetilde{\mathbf{G}}^{\rm t}$,  
\begin{align}
    \widetilde{\mathbf{G}}^{\rm t}(\mathbf{x}) = \mathbf{G}^{\rm s} (\mathcal{W}_{\rm s}^{-1}(\mathcal{W}_{\rm t}(\mathbf{x}))),\label{warp_function}
\end{align}
where $\mathcal{W}_{\rm s},\mathcal{W}_{\rm t}:\mathds{R}^2\rightarrow \mathds{R}^2$ are the warping functions that transform a point in the source and target image $\mathbf{x}$ to the UV coordinate of the body. The pseudo target garment label is incomplete because the body silhouette is a subset of the dressed body silhouette. Note that this first step, \textit{i.e.,} producing $\widetilde{\mathbf{G}}^{\rm t}$ by warping, only applies in the inference time, while in training time, we synthetically create the incomplete pseudo garment labels $\widetilde{\mathbf{G}}^{\rm t}$ by removing the outside region of the body silhouette from the ground truth $\mathbf{G}_{\rm gt}^{\rm t}$ and further removing some parts using random binary patches.
\begin{figure}
	\begin{center}
    \includegraphics[width=1\linewidth]{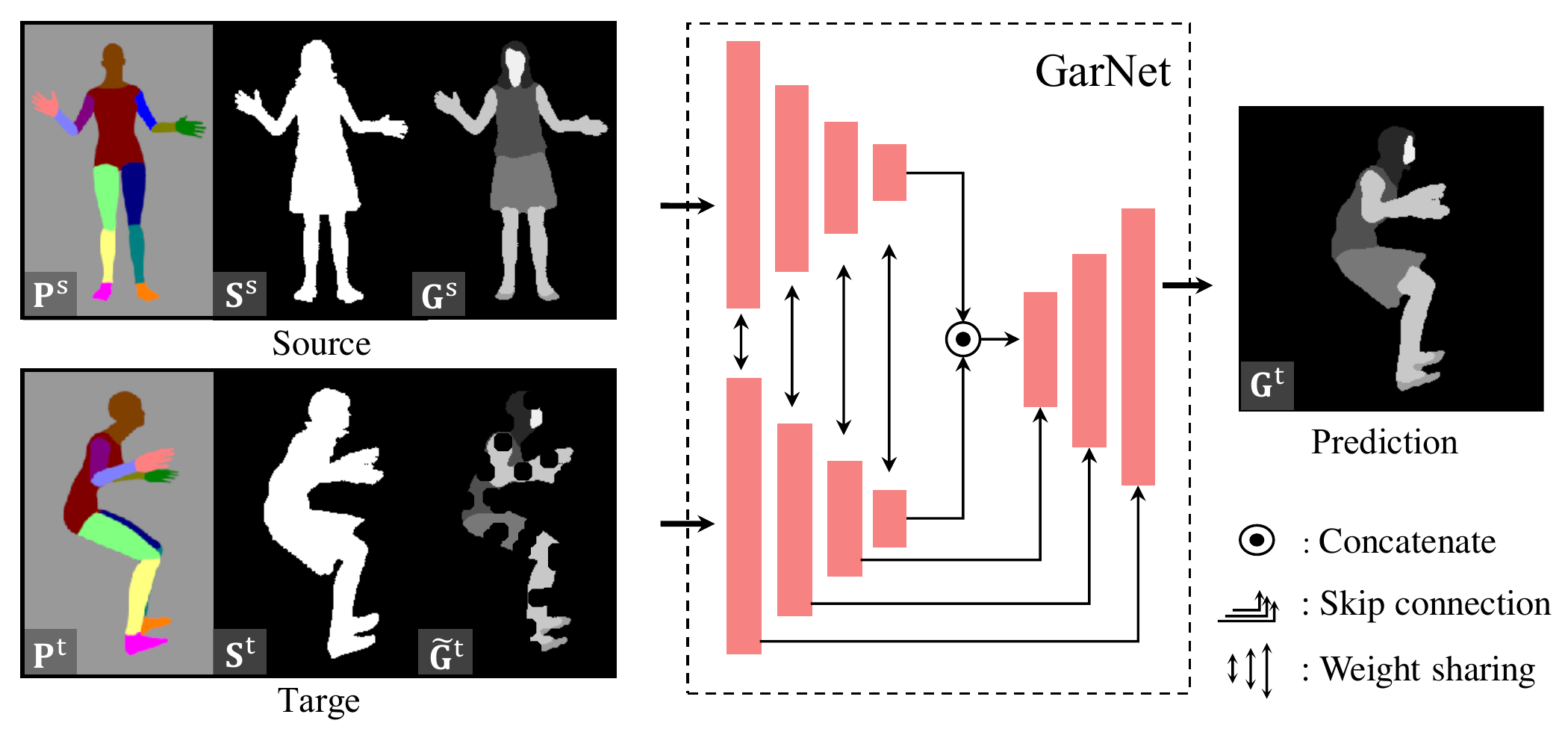}
	\end{center}	
	\vspace{-.52cm}
	\caption{\small \emph{GarNet} predicts the garment labels in the target pose. }
	\label{fig:SNet}
	\vspace{-2mm}
\end{figure}

Second, given the input triplet and the predicted target silhouette, we complete the full target garment labels $\mathbf{\mathbf{G}}^{\rm t}$:
\begin{align}
\mathbf{G}^{\rm t}=f^{\rm Gar}(\widetilde{\mathbf{G}}^{\rm t}|\mathbf{P}^{\rm t}, \mathbf{S}^{\rm t}, \{\mathbf{P}^{\rm s},\mathbf{S}^{\rm s},\mathbf{G}^{\rm s}\}).
\end{align}
We design a neural network called \emph{GarNet} to learn the target garment label completion. It consists of a Siamese encoder and a decoder, as shown in Fig.~\ref{fig:SNet}.
The Siamese encoder encodes the spatial relationship from both source and target triplets. A decoder completes the garment labels by classifying every pixel in the target silhouette. Similar to \emph{SilNet}, we use skip connections to facilitate the target feature transform. 
We train \emph{GarNet} by minimizing the following loss:
\begin{align}
L_{\rm Gar}=\left\|\mathbf{G}^{\rm t}-\mathbf{G}^{\rm t}_{\rm gt}\right\|_1.
\end{align}
$f^{\rm Gar}$ does not take as input the source image $\mathbf{I}^{\rm s}$ where using synthetic data does not introduce the domain gap.

\subsubsection{Foreground Rendering} \label{rendernet}
We synthesize the foreground person image in a target pose given the predicted target garment label and the source image triplet: source image, silhouette, and garment label. Similar to the garment label completion in Sec.~\ref{garnet}, we generate the pseudo target image $\widetilde{\mathbf{I}}^{\rm t}$ and its silhouette $\widetilde{\mathbf{S}}^{\rm t}$ using the UV coordinate transformation of $\mathcal{W}_{\rm s}$ and $\mathcal{W}_{\rm t}$ in inference time, while synthetically create the incomplete $\widetilde{\mathbf{I}}^{\rm t}$ and $\widetilde{\mathbf{S}}^{\rm t}$ from the ground truth ${\mathbf{I}}^{\rm t}_{\rm gt}$ and ${\mathbf{S}}^{\rm t}_{\rm gt}$ in training time. 

\begin{figure}
	\begin{center}
    \includegraphics[width=1\linewidth]{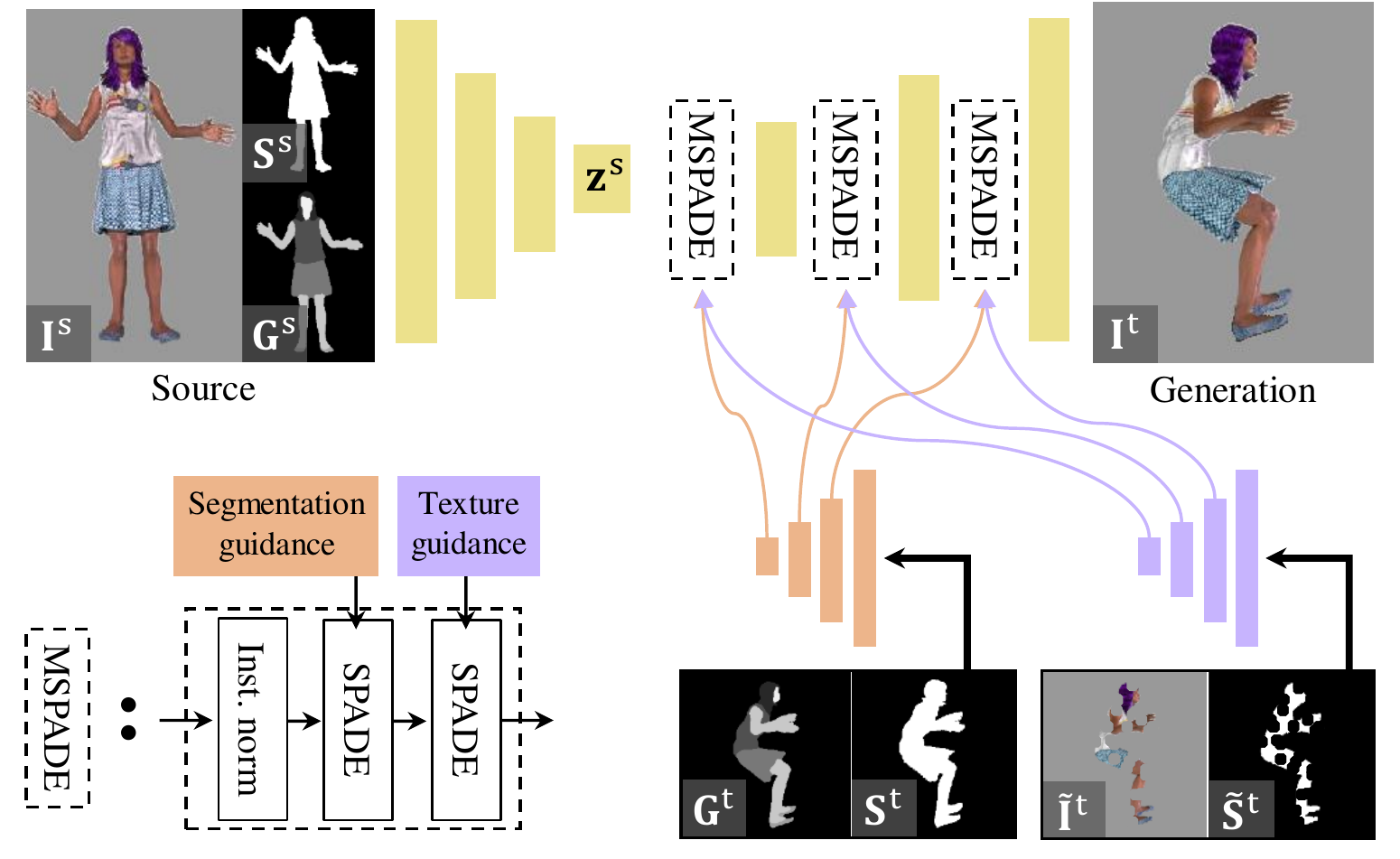}
	\end{center}	
	\vspace{-.7cm}
	\caption{\small \emph{RenderNet} synthesizes the image of a person in the target pose.}
	\vspace{-.3cm}
	\label{fig:INet}
\end{figure}

We learn a function that can render the full target foreground image:
\begin{align}
\mathbf{I}^{\rm t}=f^{\rm Render}(\widetilde{\mathbf{I}}^{\rm t},\widetilde{\mathbf{S}}^{\rm t} | \mathbf{S}^{\rm t}, \mathbf{G}^{\rm t} ,\{\mathbf{I}^{\rm s}, \mathbf{S}^{\rm s},\mathbf{G}^{\rm s}\}).
\end{align}
We design a neural network called \emph{RenderNet} to learn this function. 
As shown in Fig.~\ref{fig:INet}, \emph{RenderNet} encodes the spatial relation $\mathbf{z}^{\rm s}$ of the source image triplet, and mixes the latent representations from the target. We use two encoders to extract the features of the target garment label $\mathbf{G}^{\rm t}$ and pseudo target image $\widetilde{\mathbf{I}}^{\rm t}$ where $\mathbf{S}^{\rm t}$ and $\widetilde{\mathbf{S}}^{\rm t}$ are combined with them.  We condition these features at each level of the decoder using spatially adaptive normalization blocks~\cite{park2019semantic,mallya2020world} to guide the network to be aware of the subject's silhouette, and garment and texture style in the target pose. 

We train \emph{RenderNet} by minimizing the following loss:
%
\begin{align}
L_{Render}=L_{rec}+ \lambda_{1}L_{\text{VGG}}+\lambda_{2}L_{\text{CX}}+\lambda_{3}L_{\text{cAdv}}+\lambda_{4}L_{\text{KL}},\nonumber
\end{align}
where the weight $\lambda_i$ are empirically chosen that all the losses have comparable scale. 
\\
\textbf{Reconstruction Loss.} $L_{rec}$ measures the per-pixel errors between the synthesized image $\mathbf{I}^{\rm t}$ and the ground truth $\mathbf{I}^{\rm t}_{\rm gt}$: $L_{1}=\left\|\mathbf{I}^{\rm t}-\mathbf{I}_{\rm gt}^{\rm t}\right\|_{1}$. 
\\
\textbf{VGG Loss.} Beyond the low-level constraints in the RGB space, $L_{\text{VGG}}$ measures the image similarity in the VGG feature space~\cite{johnson2016perceptual} which is effective in generating natural and smooth person image proven by existing works~\cite{men2020controllable,esser2018variational,tang2020xinggan}:
%
%
$L_{\text{VGG}}=\sum_{i=1}^4\left\|\text{VGG}_{i}(\mathbf{I}^{\rm t})-\text{VGG}_{i}(\mathbf{I}_{\rm gt}^{\rm t})\right\|_{1}$, where $\text{VGG}_i(\cdot)$ maps an image to the activation of the conv-i-2 layer of VGG-16 network~\cite{simonyan2014very}.
\\
\textbf{Contextual Loss.} $L_{\text{CX}}$ measures the similarity of two set of features considering global image context:
$L_{\text{CX}}=-\operatorname{log}(g(\text{VGG}_{3}(\mathbf{I}^{\rm t}),\text{VGG}_{3}(\mathbf{I}^{\rm t}_{\rm gt})), \label{cont_loss}$ where $g(\cdot,\cdot)\in[0,1]$ denotes the similarity metric of the matched features based on the normalized cosine distance~\cite{mechrez2018contextual}. Existing work~\cite{men2020controllable} proved that combining $L_{\text{CX}}$ with $L_{\text{VGG}}$ further helps to preserve the style patterns in the generated image in a semantically meaningful way, \textit{i.e.,} less distorted facial structure.
\textbf{Adversarial loss.} We employ the conditional adversarial loss $L_{\text{cAdv}}$~\cite{mirza2014conditional} with a discriminator conditioned on garment labels to classify the synthesized image into real or fake, \textit{i.e.,}  $\{\mathbf{I}^{\rm t}_{\rm gt},\mathbf{G}^{\rm t}_{\rm gt}\}$ is real and $\{\mathbf{I}^{\rm t},\mathbf{G}^{t}_{\rm gt}\}$ is fake. Here, we use the PatchGAN discriminator~\cite{isola2017image}. 
\\
\textbf{KL divergence.} $L_{\text{KL}}$ is to enforce the latent space $\mathbf{z}^{\rm s}$ to be close to a standard normal distribution~\cite{kullback1951information,kingma2013auto}.

\begin{figure*}
	\begin{center}
    \includegraphics[width=1\linewidth]{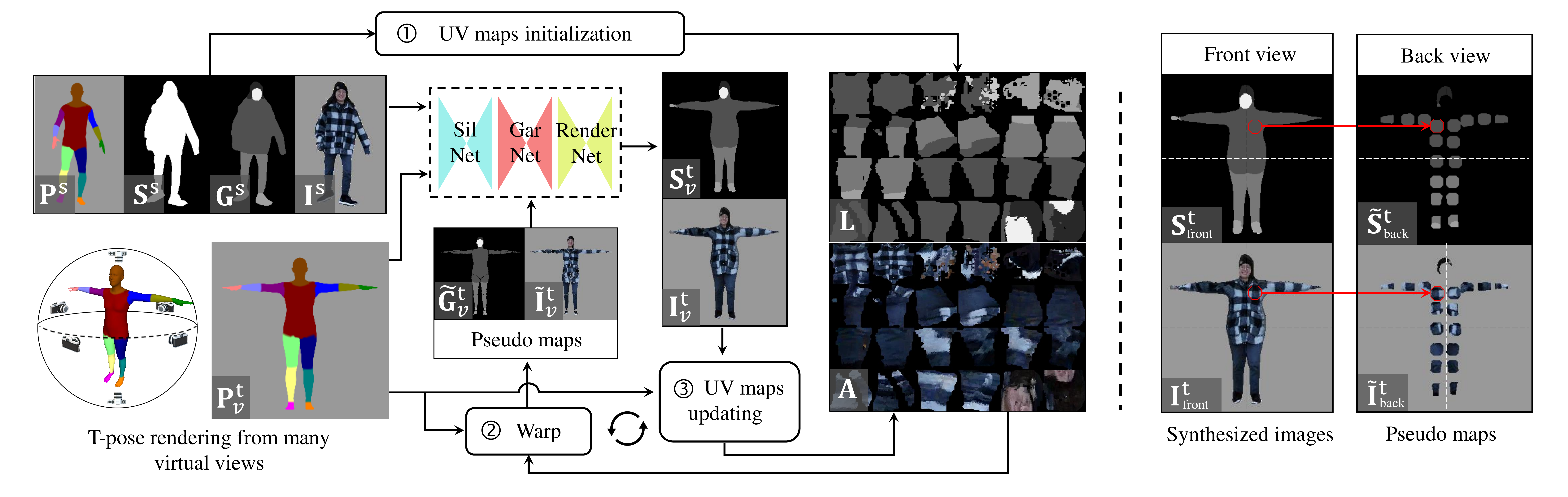}
	\end{center}	
	\vspace{-.8cm}
	\caption{\small We reconstruct the complete UV maps of the garment labels and textures, \textit{i.e.,} $\mathbf{L}$ and $\mathbf{A}$, in an incremental manner. (Left) We first initialize these maps by warping the pixels in the source image, \textit{i.e.,} $\mathbf{I}^{\rm s}$ and $\mathbf{S}^{\rm s}$, to the UV maps. We further update the UV maps by combining the synthesized images of a person in a T pose captured from six  virtual views. For each virtual view $v$, we create the pseudo images, \textit{i.e.,} $\widetilde{\mathbf{G}}^{\rm t}_{v}$ and $\widetilde{\mathbf{I}}^{\rm t}_{v}$, from the previously updated UV maps. (Right) Only for the back view, we construct $\widetilde{\mathbf{G}}^{\rm t}_{v}$ and $\widetilde{\mathbf{I}}^{\rm t}_{v}$ by sampling the patches from the synthesized images in the frontal view with the front-back symmetry assumption where the face regions are removed.}
	\label{fig:uv_maps}
\end{figure*}
\subsection{Consistent Human Animation Creation}\label{human_ani}

With the learned pose transfer network, it is possible to generate the shape and appearance given a target pose map at each time instant. However, it makes independent prediction for each pose, which leads to unrealistic jittery animation. Instead, we build a unified representation of appearance and its labels that provide a consistent guidance across different poses, which enforces the network to predict temporally coherent appearance and shape.

We construct the garment labels $\mathbf{L}$ and textures $\mathbf{A}$ that remain constant in UV coordinates by warping the garment label and appearance of an image, \textit{i.e.,} $\mathbf{L}(\mathbf{x}) = \mathbf{G}(\mathcal{W}^{-1}(\mathbf{x}))$ and $\mathbf{A}(\mathbf{x}) = \mathbf{I}(\mathcal{W}^{-1}(\mathbf{x}))$. These UV representations ($\mathbf{L}$ and $\mathbf{A}$) cannot be completed from a single view input image because of occlusion. To complete the UV representations, we use the multiview images synthesized from the rendered 3D human model of which texture is predicted by the learned pose transfer network. 
This set of generated images are used to incrementally complete the UV representations as shown in Fig.~\ref{fig:uv_maps}-(left). 

In practice, we generate multiview images by synthesizing the SMPL model at the T pose from six views: front, back, left, right, top and bottom views. We assume that the source image is taken from the frontal view. The back view is generated by applying front-back symmetry assumption~\cite{natsume2019siclope,gabeur2019moulding,weng2019photo} as shown in Fig.~\ref{fig:uv_maps}-(right). 

In the inference phase, this unified UV representation allows us to consistently generate the pseudo garment labels $\widetilde{\mathbf{G}}^{\rm t}(\mathbf{x}) = \mathbf{L}(\mathcal{W}_{\rm t}(\mathbf{x}))$ and appearance $\widetilde{\mathbf{I}}^{\rm t}(\mathbf{x}) = \mathbf{A}(\mathcal{W}_{\rm t}(\mathbf{x}))$ given a target pose by transforming the SMPL T-pose to the target pose. This pseudo representations provide an incomplete yet strong guidance to the pose transfer network to complete the target foreground.

In order to have both foreground and background in the animation, we segment the foreground from the source image using $\mathbf{S}^{\rm s}$ and apply an inpainting  method~\cite{yu2019free} to the background. 
We then composite our synthesized human animation with the background. 

\section{Implementation Details}

We train the proposed \emph{SilNet}, \emph{GarNet}, and \emph{RenderNet} separately in a fully supervised way using only 3D people synthetic dataset~\cite{pumarola20193dpeople} which is described in Sec~\ref{pose_transfer}. For training, we set the parameters of $\lambda_{1}=0.5$, $\lambda_{2}=0.1$, $\lambda_{3}=0.01$, $\lambda_{4}=10$
and use the Adam optimizer~\cite{KingmaBa2014} ($lr=1\times10^{-3}$ and $\beta=0.5$).
After training, no further fine-tuning on the testing scene is required. For the pose map $\mathbf{P}$ and garment label map $\mathbf{S}$, we convert them to rgb and gray scale images for the network input. 

In inference time, we obtain $\mathbf{S}^{\rm s}$ and $\mathbf{G}^{\rm s}$ using person segmentation~\cite{yolact-iccv2019} and fashion segmentation~\cite{gong2017look}. For $\mathbf{P}^{\rm s}$, we fit a 3D body model~\cite{SMPL:2015} to an image using recent pose estimator~\cite{kolotouros2019learning} and render the parts label onto the image where we follow the same color coding as synthetic data~\cite{pumarola20193dpeople}. We generate a sequence of body poses $\{\mathbf{P}_{i}^{\rm t}\}_{i=1}^{N}$ by animating the 3D body model using recent motion archive~\cite{AMASS:ICCV:2019}, where we represent the \textit{z}-directional motion as scale variation~\cite{kanazawa2018end} with weak-perspective camera projection, and rendering the pose map from each body pose similarly to $\mathbf{P}^{\rm s}$. The image resolution is $256\times256$, and UV maps are $512\times 768$.

\section{Experiments}\label{sec:exp}
In order to evaluate our approach, we collect eight sequences of the subjects in various clothing
and motions from existing works~\cite{shimada2020physcap,yoon2020dynamic,liu2019liquid,alldieck2018video,habermann2020deepcap}
and capture two more sequences which include a person with more complex clothing style and motion than others.
Each sequence contains 50 to 500 frames. We use one frame in the sequence as source image and estimated body poses from the rest of frames using a pose estimator~\cite{kolotouros2019learning} as a target pose sequence.

{
\renewcommand{\tabcolsep}{5pt}

\begin{table*}[t]
\centering
\small
\vspace{-1mm}
\begin{tabular}{l|c|c|c|c|c|c|c|c|c|c|c}
\hline
 & \scriptsize{Maskman}  & \scriptsize{Rainbow}  & \scriptsize{RoM1}  & \scriptsize{RoM2} &  \scriptsize{Jumping} & \scriptsize{Kicking} & \scriptsize{Onepiece}  & \scriptsize{Checker} & \scriptsize{Rotation1}& \scriptsize{Rotation2} & \scriptsize{Average}   \\
\hline
\scriptsize{PG (3P)} &\scriptsize{3.45 / 4.46}  &\scriptsize{3.28 / 4.58}   &\scriptsize{3.38 / 4.37}    &  \scriptsize{3.30 / 4.57}&\scriptsize{3.87 / 4.68}  &  \scriptsize{2.98 / 4.36}&\scriptsize{3.26 / 4.63}&  \scriptsize{3.03 / 4.45}&\scriptsize{3.67 / 4.41}&\scriptsize{2.95 / 4.06}&\scriptsize{2.93 / 4.45}     \\

\scriptsize{SGAN (3P)}&\scriptsize{1.93 / 2.97}  &\scriptsize{1.61 / 3.04}   &\scriptsize{1.62 / 2.94}  &\scriptsize{1.60 / 3.02}  &\scriptsize{2.27 / 3.12}  &\scriptsize{1.56 / 2.98}  &\scriptsize{1.82 / 3.06}  &\scriptsize{1.53 / 3.03}  & \scriptsize{1.56 / 3.01} & \scriptsize{1.65 / 2.84} & \scriptsize{1.71 / 3.00} \\ 

\scriptsize{PPA (3P)}&\scriptsize{1.88 / 2.89}  &\scriptsize{1.62 / 2.95}   &\scriptsize{1.40 / 2.82}  &\scriptsize{1.66 / 3.00}  &\scriptsize{2.43 / 3.03}&\scriptsize{1.33 / 2.88}  &\scriptsize{1.86 / 2.95}  &\scriptsize{1.49 / 2.88}  & \scriptsize{1.38 / 2.84} & \scriptsize{1.40 / 2.69} & \scriptsize{1.64 / 2.89} \\ 

\scriptsize{GFLA (3P)}&\scriptsize{1.90 / 2.92}  &\scriptsize{1.59 / 3.05} &\scriptsize{1.53 / 2.91}  &\scriptsize{1.71 / 2.95}  &  \scriptsize{2.11 / 3.06}  &\scriptsize{1.42 / 2.97}  &\scriptsize{1.60 / 3.03}  &\scriptsize{1.64 / 2.97}   &\scriptsize{1.64 / 2.93}&\scriptsize{1.65 / 2.80}&\scriptsize{1.68 / 2.96} \\ 

\scriptsize{NHHR (3P)}&\scriptsize{1.65 / 2.81}  &\scriptsize{1.61 / 2.94}  & \scriptsize{1.49 / 2.80}& \scriptsize{1.41 / 2.88} & \scriptsize{1.99 / 3.01} &   \scriptsize{\textbf{1.00} / 2.85} & \scriptsize{1.78 / 2.98} & \scriptsize{1.65 / 2.94} & \scriptsize{1.39 / 2.88} & \scriptsize{1.67 / 2.75}  & \scriptsize{1.56 / 2.88}   \\ 
\hline

\scriptsize{LWG}&\scriptsize{2.66 / 3.54}  &\scriptsize{2.16 / 3.57}  & \scriptsize{2.17 / 3.49} & \scriptsize{2.24 / 3.73} &   \scriptsize{4.11 / 3.49} & \scriptsize{2.42 / 3.57} & \scriptsize{2.31 / 3.65} & \scriptsize{2.40 / 3.57} & \scriptsize{2.14 / 3.64}  & \scriptsize{2.20 / 3.48}  & \scriptsize{2.48 / 3.57} \\ 


\hline
\scriptsize{Ours}&  \scriptsize{\textbf{1.54} / \textbf{2.27}} & \scriptsize{\textbf{1.24} / \textbf{2.38}}& \scriptsize{\textbf{1.25} / \textbf{2.24}} & \scriptsize{\textbf{1.38} / \textbf{2.36}}  &\scriptsize{\textbf{1.87} / \textbf{2.53}}  &\scriptsize{1.08 / \textbf{2.19}}  &\scriptsize{\textbf{1.23} / \textbf{2.32}}  &\scriptsize{\textbf{1.09} / \textbf{2.24}}  &\scriptsize{\textbf{1.00} / \textbf{2.19}}&\scriptsize{\textbf{1.12} / \textbf{2.16}}&\scriptsize{\textbf{1.28} / \textbf{2.29}}\\ 

\hline
\end{tabular}
\vskip-7pt
\caption{\small Quantitative results with LPIPS (left, scale: $\times10^{\shortminus1}$) and CS (right) where the lower is the better.}
\label{table:eval}
\vskip-6pt
\end{table*}

}
\begin{figure*}
	\begin{center}
	\vspace{-1mm}
\includegraphics[width=1.0\linewidth]{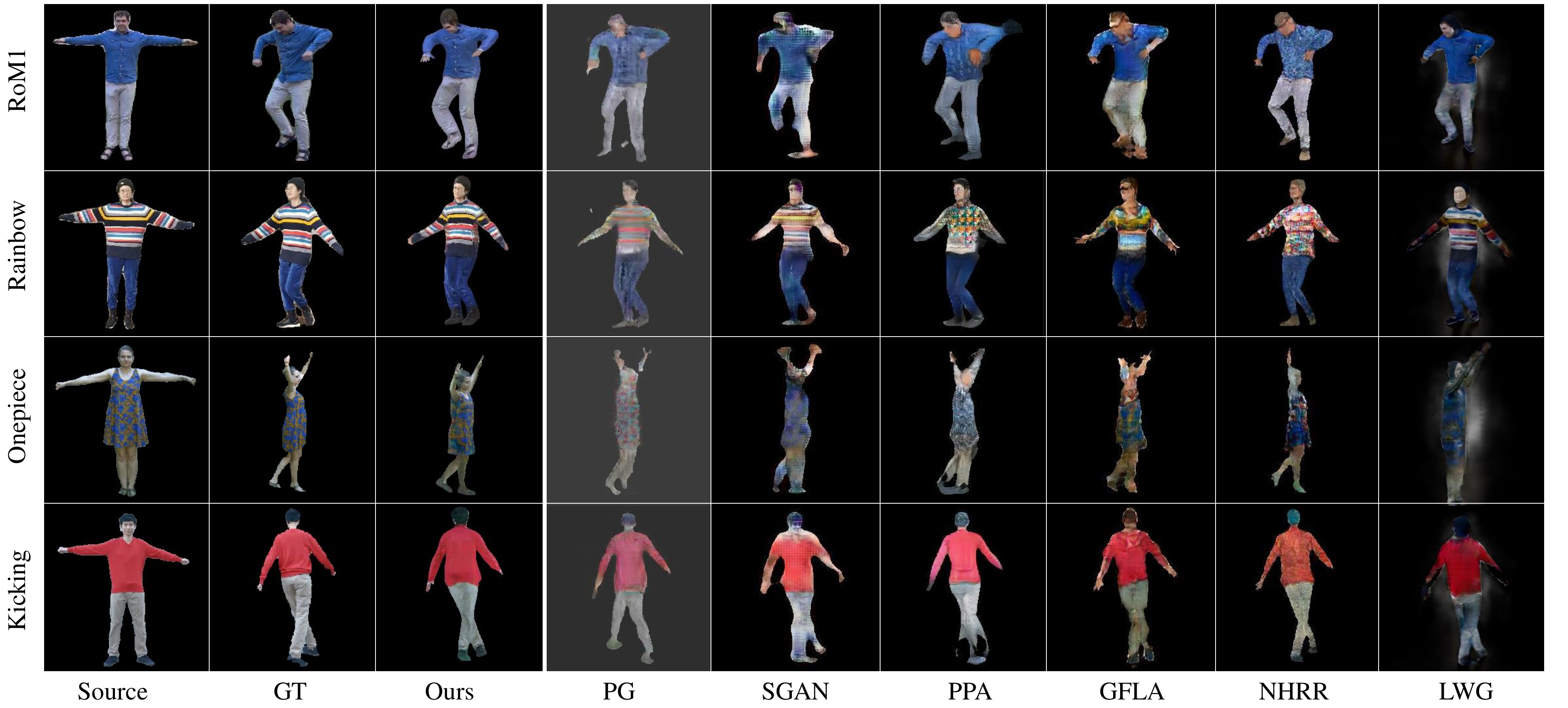}
	\end{center}	
	\vspace{-.7cm}
	\caption{\small Qualitative comparisons of our approach with other baseline methods. See our supplementary video for more  results.  
	}
	\vspace{-5mm}
	\label{qualitative}
\end{figure*}

\noindent\textbf{Baseline.} We compare our method with related works including \textit{PG}~\cite{ma2017pose}, \textit{SGAN}~\cite{tang2020multi}, \textit{PPA}~\cite{zhu2019progressive}, \textit{GFLA}~\cite{ren2020deep}, \textit{NHRR}~\cite{sarkarneural}, \textit{LWG}~\cite{liu2019liquid}.
Note that all these methods except LWG are not designed to handle background.
We compare all the methods on foreground synthesis and conduct an additional comparison with LWG on the full image synthesis including both foreground and background. 
For a fair comparison, we train all the methods except LWG on 3D people dataset~\cite{pumarola20193dpeople}. 
For training, LWG requires a SMPL model which is not provided by the 3D people dataset. 
Since registering a SMPL model to each 3D model in the 3D people dataset may introduce fitting error, we use the pretrained model provided by the authors, which are trained on the iPER dataset~\cite{liu2019liquid}. 
We also evaluate the methods with the pretrained models provided by the authors, which were trained on the Deep Fashion dataset~\cite{liu2016deepfashion} (see the supplementary material). 
In addition, we provide a qualitative comparison with \textit{Photo Wake-Up}~\cite{weng2019photo} which reconstructs a textured animatable 3D model from a single image.

\begin{table*}[t]
\centering
\small
\vspace{-1mm}
\begin{tabular}{|l||c|c|c||c|c||c|c|c|c||c|c||c|c||c|}
\hline
&\scriptsize{R}  & \scriptsize{GR}  & \scriptsize{SR}& \scriptsize{SGR-$\mathbf{M}^{\rm s}$} &  \scriptsize{SGR-$\mathbf{S}^{\rm s}$} &  \scriptsize{SGR-$\mathbf{z}^{\rm s}$} &\scriptsize{SGR-$L_{\rm KL}$}  &\scriptsize{SGR-$\widetilde{\mathbf{I}}^{\rm t}$}&\scriptsize{SGR-$\mathbf{A}$} & \scriptsize{SGR+2view} & \scriptsize{SGR+4view} &\scriptsize{SGR (\textbf{full})}   \\
\hline
\scriptsize{LPIPS} &\scriptsize{0.148}  &\scriptsize{0.147}   &\scriptsize{0.133}     &  \scriptsize{0.135}&\scriptsize{0.141}&\scriptsize{0.132}&\scriptsize{0.130}&\scriptsize{0.162}&\scriptsize{0.136}
&  \scriptsize{0.125}&\scriptsize{\textbf{0.124}}&\scriptsize{0.128} \\
\hline
\scriptsize{CS} &\scriptsize{2.354}  &\scriptsize{2.352}   &\scriptsize{2.311} &  \scriptsize{2.322}&\scriptsize{2.342} &\scriptsize{2.308}&\scriptsize{2.301}&\scriptsize{2.335}&\scriptsize{2.310} &  \scriptsize{2.281}&\scriptsize{\textbf{2.274}}&\scriptsize{2.292}     \\
\hline
\end{tabular}
\vskip-7pt
\caption{\small Quantitative results of our ablation study. We denote our complete model with a single image as input as SGR(\textbf{full}).}
\label{ablation_study}
\vskip-15pt
\end{table*}

\begin{figure}
	\begin{center}
	\vspace{1.25mm}
    \includegraphics[width=1\linewidth]{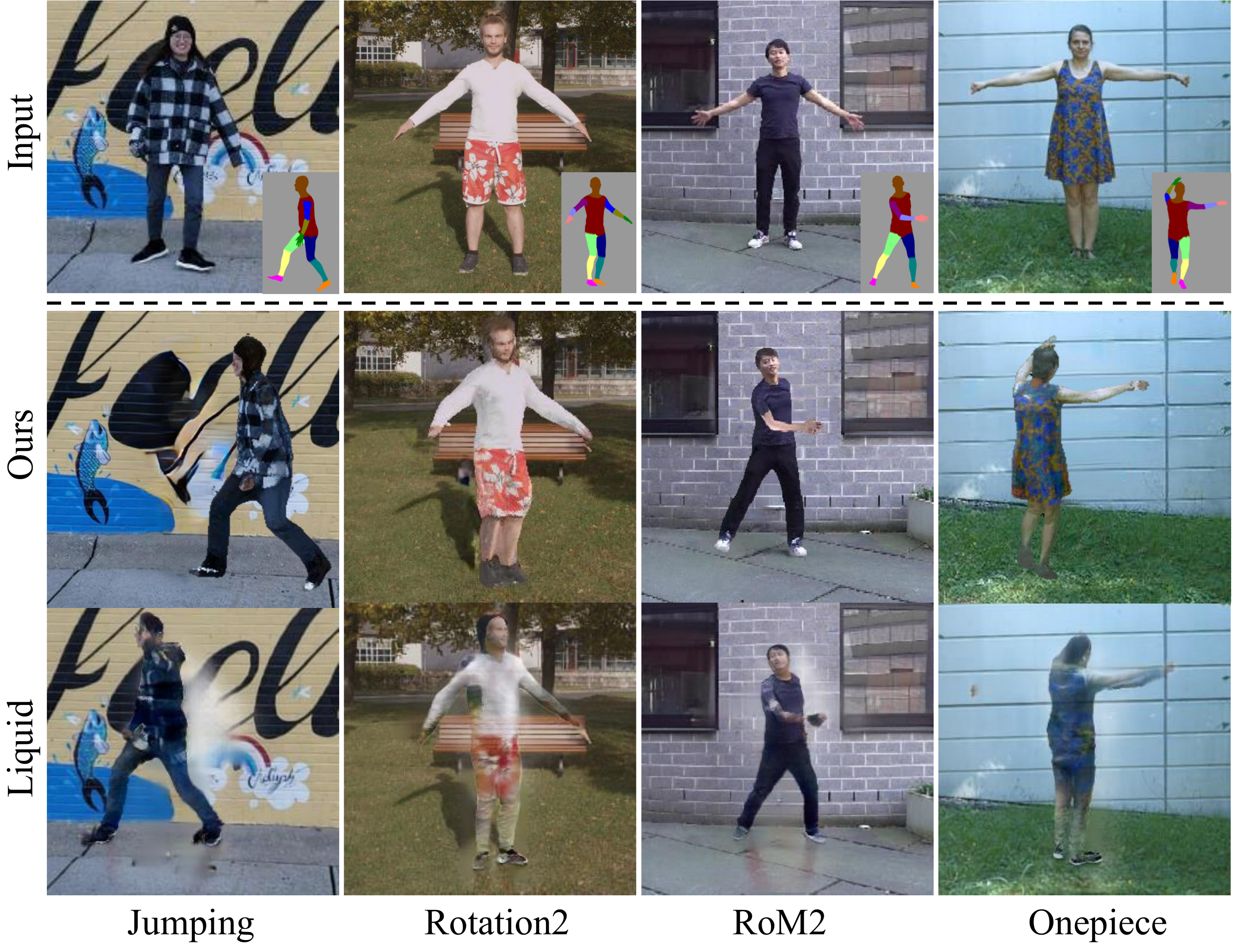}
	\end{center}	
	\vspace{-.5cm}
	\caption{\small Qualitative comparison with LWG on the input images with background. The target pose is shown as inset. See our supplementary video for more results.  
	}
	\vspace{-.5cm}
	\label{res_w_back}
\end{figure}
\begin{figure}
	\begin{center}
    \includegraphics[width=1.06\linewidth]{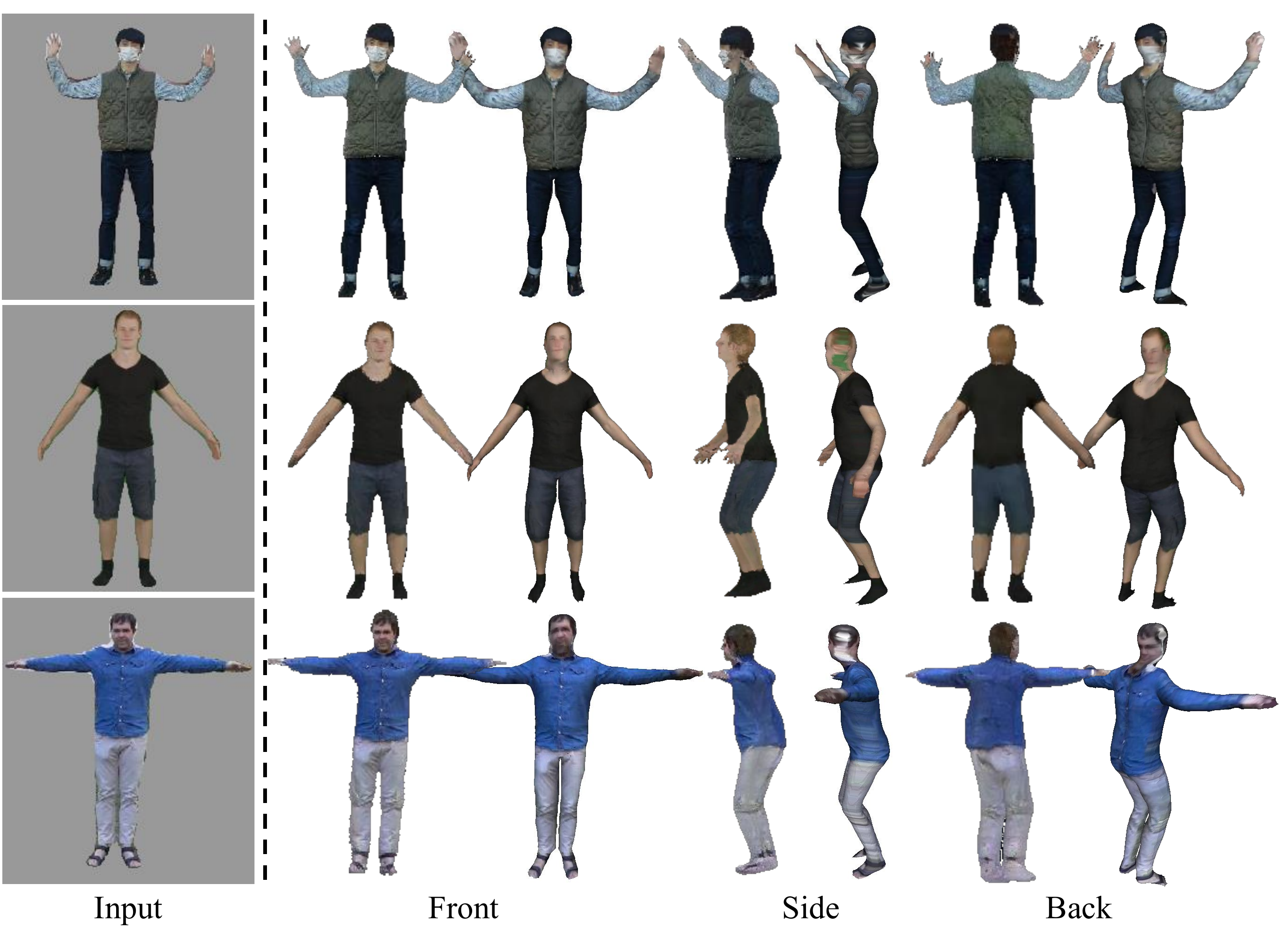}
	\end{center}	
	\vspace{-.5cm}
	\caption{\small Qualitative comparison of ours (left) with Photo Wake-Up (right). 
	}
	\label{photo-wakeup}
\end{figure}


\subsection{Comparisons}
\paragraph{Qualitative Comparisons.} 
We show the qualitative comparison with the baselines on the foreground synthesis in Fig.~\ref{qualitative}.
Note that for the results in Fig.~\ref{qualitative}, all methods are trained on 3D people datasets (see the results of the baselines trained on the Deep Fashion dataset in the supplementary video). 
Our method significantly outperforms other baselines as preserving the facial identity, body shape and texture patterns of clothes over all the subjects with various challenging poses.
Furthermore, compared to the baseline methods, our method generalizes better on the real data and achieves more realistic results that are close to the ground truth, although only synthetic data is used for training. 
We conduct a comparison with LWG on the full image synthesis, where our method can synthesize higher quality foreground as well as background, as shown in Fig.~\ref{res_w_back}. Compared to Photo Wake-Up in Fig.~\ref{photo-wakeup}, we can render the better textures on the right and back side of the person.

\begin{figure}
	\begin{center}
	\vspace{-0.5mm}
    \includegraphics[width=1.02\linewidth]{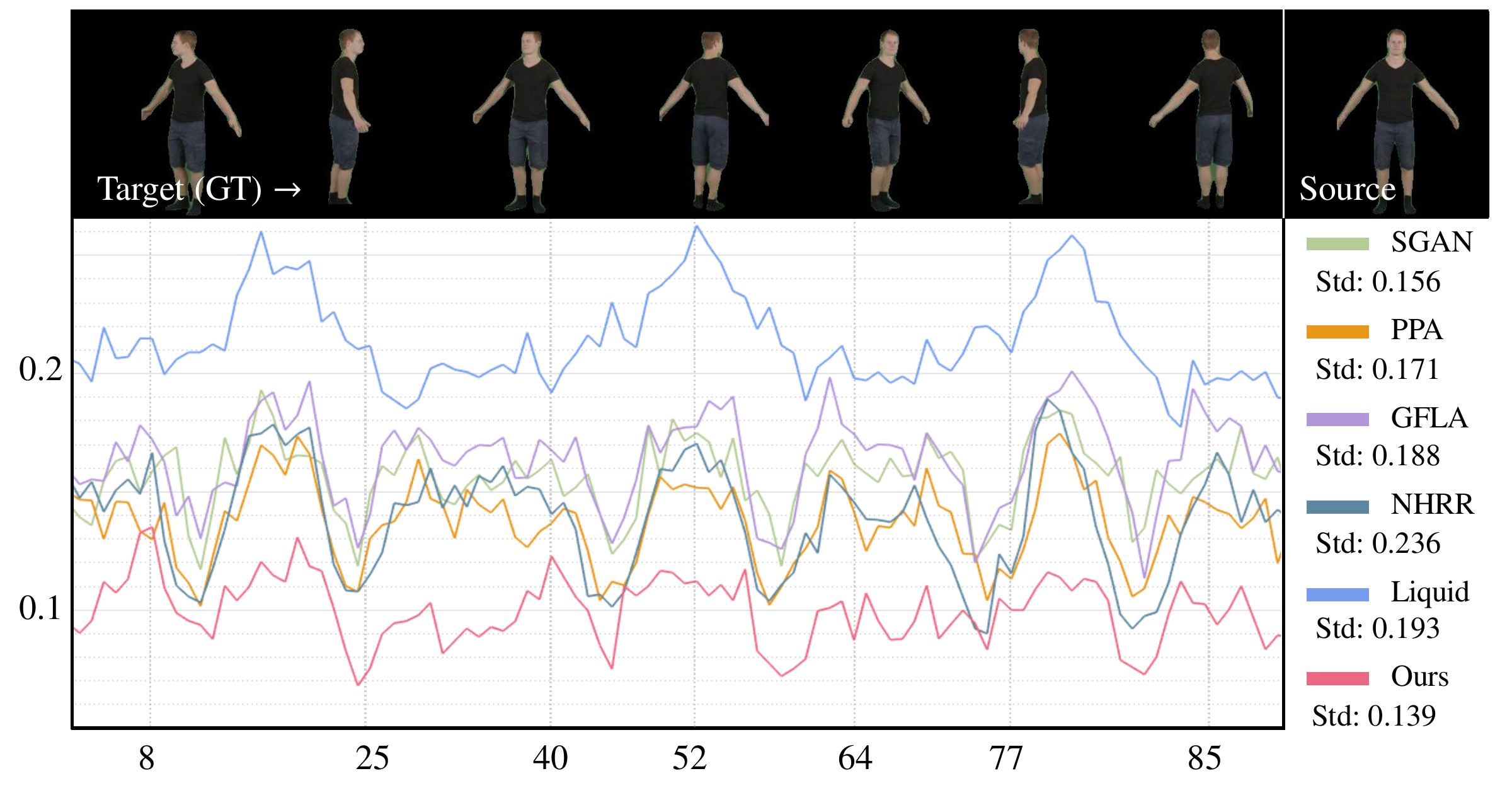}
	\end{center}	
	\vspace{-.7cm}
	\caption{\small The accuracy graph for the entire frames of a video. \textit{x}-axis and \textit{y}-axis represent time instance and LPIPS, respectively.} 
	\label{qual_temporal}
\end{figure}

\paragraph{Quantitative Comparisons.} 
We measure the quality on testing results with two metrics: LPIPS~\cite{zhang2018unreasonable} and CS~\cite{mechrez2018contextual} where both metrics measure the similarity of the generated image with ground truth based on the deep features, and CS can handle the non-aligned two images. As shown in Table~\ref{table:eval}, our method outperforms all baseline methods over almost all the sequences in LPIPS and CS. In Kicking, our method performs the second best in LPIPS metric mainly due to the misalignment with the ground truth originated from the pose estimation error. In Fig.~\ref{qual_temporal}, we measure temporal stability of the synthesized animations with the standard deviation of the LPIPS scores with respect to all the frames, where our results show the best temporal stability.


\subsection{Ablation Study}
We study the importance of each module in our pose transfer pipeline where we term ``S'', ``G'', and ``R'' as \emph{SilNet}, \emph{GarNet}, and \emph{RenderNet}, and our full model as \textit{SGR}.

\begin{figure}
	\begin{center}
    \includegraphics[width=1\linewidth]{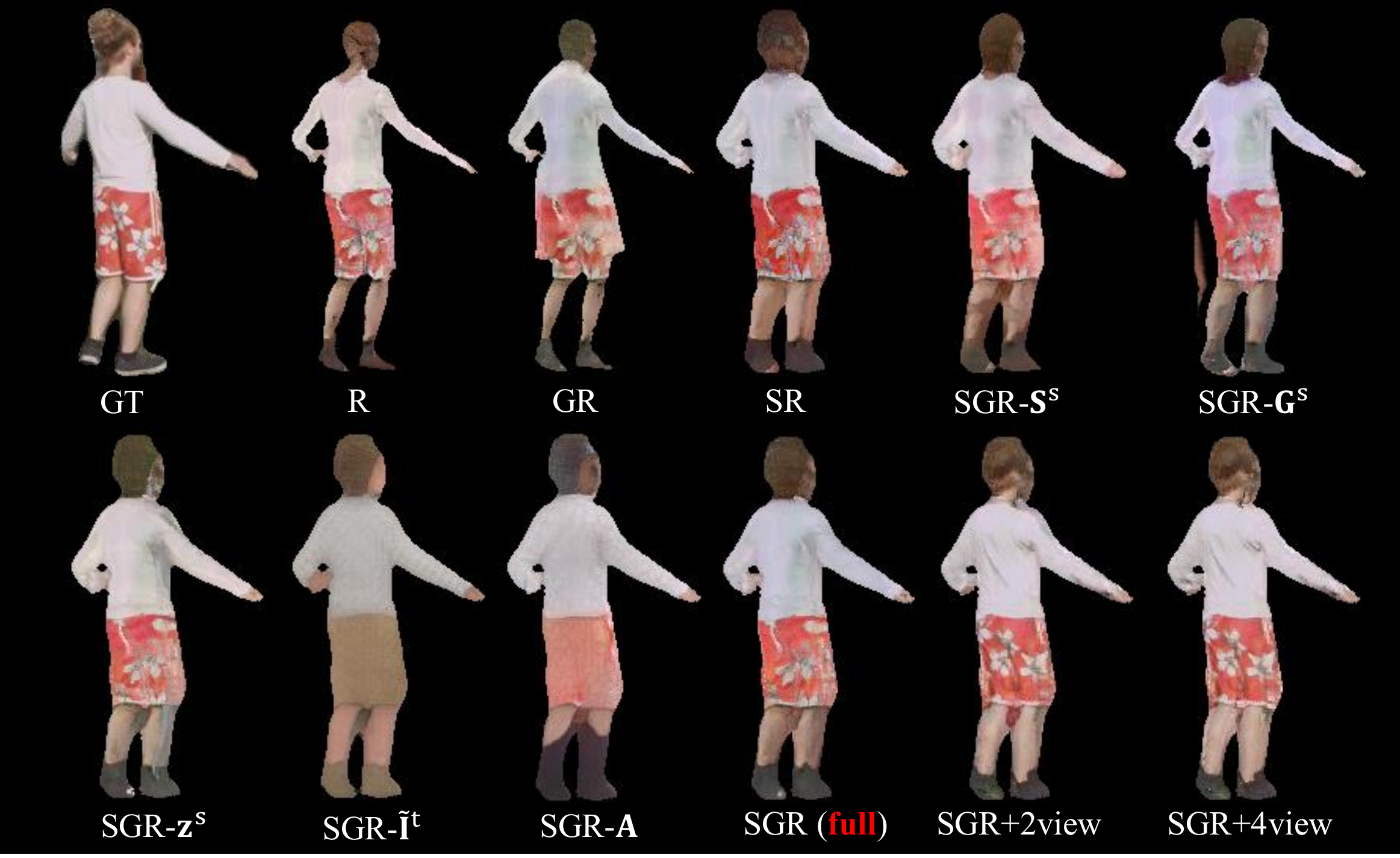}
	\end{center}	
	\vspace{-.5cm}
	\caption{\small Qualitative results of our ablation study. See our supplementary video for more results.  
	}
	\label{qual_ablation}
\end{figure}

\noindent1) We analyze the effectiveness of our modular network by removing each from SGR where the intermediate results are also removed from the entire pipeline: \textit{R}, \textit{SR}, and \textit{GR}. 
\\
2) We evaluate the impact of using silhouette mask and garment label from the source by removing each of them from the entire pipeline: \textit{SGR}-$\mathbf{S}^{\rm s}$ and \textit{SGR}-$\mathbf{M}^{\rm s}$. 
\\
3) We investigate the improvement factor on the \emph{RenderNet}: \textit{SGR}-$\mathbf{z}^{\rm s}$, \textit{SGR}-$\widetilde{\mathbf{I}}^{\rm t}$,  and \textit{SGR}-$L_{\text{KL}}$. For \textit{SGR}-$L_{\text{KL}}$, we represent the latent space with fully connected layers. On top of that, we investigate the impact of reconstructing a complete UV map: \textit{SGR}-$\mathbf{A}$. In this case, we create the pseudo target image $\widetilde{\mathbf{I}}^{\rm t}$ by directly warping the source image to the target.
\\
4) Finally, we show that our method is readily extendable to the multiview setting by unifying all the pixels from multiple images in the coherent UV maps. For this, we choose two or four frames from the testing videos that include salient body sides, \textit{e.g.,} front, back, right, and left: \textit{SGR}+2view and \textit{SGR}+4view.
\\
We summarize the results of our ablation study in Table~\ref{ablation_study} and the qualitative results are shown in Fig.~\ref{qual_ablation}. Separating the silhouette prediction module from rendering network brings out notable improvement, and the predicted garment labels $\mathbf{G}^{\rm t}$ further improve the results, \textit{e.g.,} clear boundary between different classes. Without the garment labels from the source $\mathbf{G}^{\rm s}$ the performance is largely degraded due to the misclassified body parts.
Conditioning the style code $\mathbf{z}^{\rm s}$ from the source improves the generation quality, \textit{e.g.,} seamless inpainting.
Conditioning the pseudo images $\widetilde{\mathbf{I}}^{\rm t}$ warped from the coherent UV maps $\mathbf{A}$ plays the key role to preserve the subject's appearance in the generated image. Leveraging multiview images better can preserve the clothing texture, \textit{e.g.,} the flower patterns in the subject's half pants.
\subsection{User Study} 
We evaluate the qualitative impact of our method 
by a user study with 25 videos where each video shows a source image and animated results. 
Four videos compare our method to LWG on the scenes with a background. 
21 videos are without background (15 of them compare our method to randomly-chosen four baselines, excluding ground truth, and 6 videos include ground truth).  
%
%
%
%
47 people participated in total. 
In $84.3\%$ and $93\%$ of the cases, our method was found to produce the most realistic animations in the settings with and without ground truth, respectively. 
Moreover, these numbers strongly correlate with the identity-preserving properties of our method. 
Finally, our technique preserves the background better compared to LGW, in the opinion of respondents ($96.8\%$ of the answers). 
%
%
%
The user study shows that our method significantly outperforms the state of the arts in terms of synthesis quality, temporal consistency and generalizability. 
Also, our results were often ranked as more realistic than the ground truth videos. The full results will be shown in the supplementary material.

\subsection{Limitations}
Our method has several limitations. Although the unified representation of appearance and its labels allow us to synthesize temporally consistent results, it prevents from generating realistic physical effects such as pose-dependent clothing secondary motion, wrinkles, shading, and view-dependent lighting. Because of non-end-to-end nature of our method, the errors from the pre-processing step, \textit{e.g.,} person and garment segmentation, and pose estimation, cannot be corrected by our pose transfer network. 
\section{Conclusion}

We introduce a new pose transfer framework to animate humans from a single image. 
%
%
We addressed the core domain gap challenge for the testing data in the wild by designing a new compositional pose transfer network that predicts silhouette, garment labels, and textures in series, which are learned from synthetic data.  
%
In inference time, we reconstruct coherent UV maps by unifying the source and synthesized images, and utilize these UV maps to guide the network to create coherent human animation.
%
%
The evaluation on diverse subjects demonstrates that our framework works well on the unseen data without any fine-tuning and preserves the identity and texture of the subject as well as background in a temporally coherent way, showing a significant improvement over the state-of-the-arts. 
%

\section*{Acknowledgement}
Christian Theobalt, Vladislav Golyanik, Kripasindhu Sarkar were supported by the ERC Consolidator Grant 4DRepLy (770784). Lingjie Liu was supported by Lise Meitner Postdoctoral Fellowship. This work is also supported by NSF CAREER IIS-1846031 and NSF CNS-1919965.



{\small
\bibliographystyle{ieee_fullname}
\bibliography{main.bbl}
}

\clearpage

\setcounter{section}{0}
\def\thesection{\Alph{section}}
\noindent This supplementary material provides additional implementation details of our compositional pose transfer network (Sec.~\ref{network}) and more results (Sec.~\ref{more_res}). In the supplementary video, we included the full results of the qualitative comparison, ablation study, more results, and the description of our overall pipeline.

  \begin{figure}
	\begin{center}
    \includegraphics[width=0.9\linewidth]{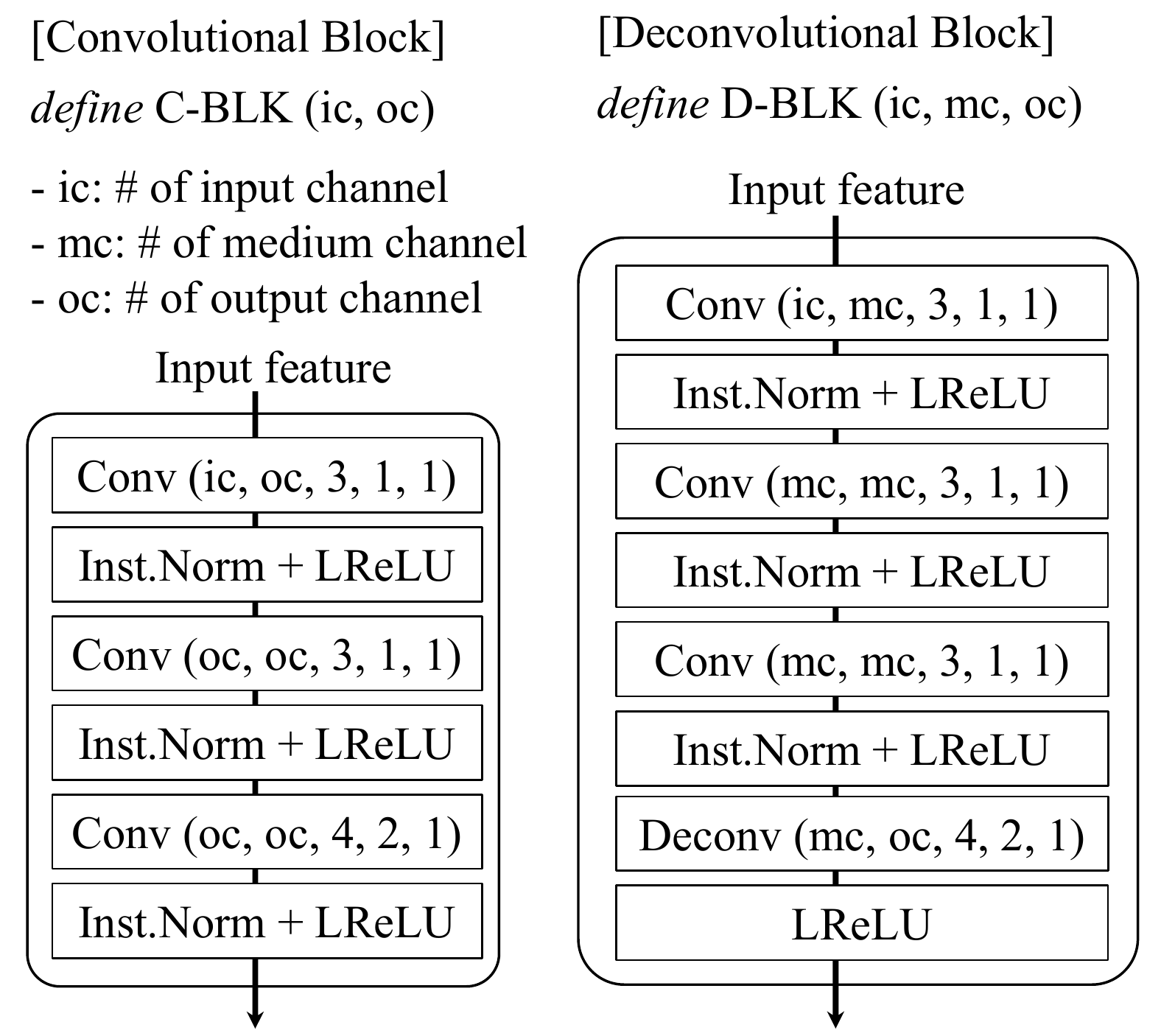}
	\end{center}	
	\vspace{-.5cm}
	\caption{\small Description of our convolutional and deconvolutional blocks. The convolutional (Conv) and deconvolutional layers (Deconv) take  parameters including the number of input channels, the number of output channels, filter size, stride, and the size of zero padding. We use 0.2 for the LeakyReLU (LReLU) coefficient.}
	\label{cov_block_network}
\end{figure}

\begin{figure}
	\begin{center}
    \includegraphics[width=0.8\linewidth]{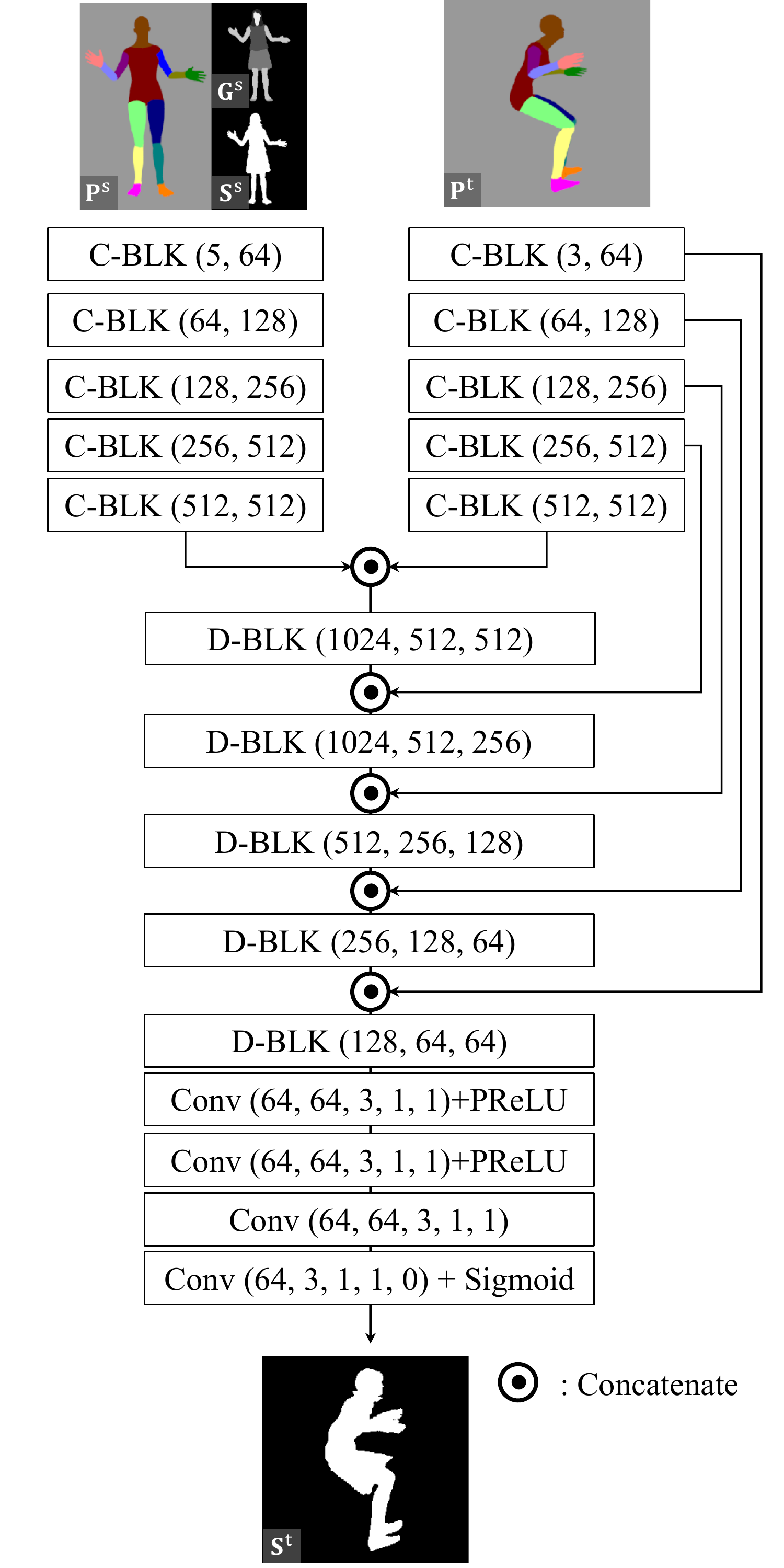}
	\end{center}	
	\vspace{-.5cm}
	\caption{\small The details of our \emph{SilNet} implementation where C-BLK and D-BLK are described in Fig.~\ref{cov_block_network}. Conv and Deconv take as input parameters of (the number of input channels, the number of output channels, filter size, stride, the size of zero padding). We use 0.2 for the LeakyReLU (LReLU) coefficient.}
	\label{silnet_network}
\end{figure}

\section{Additional Implementation Details}\label{network}
\begin{figure}
\begin{center}
\includegraphics[width=0.8\linewidth]{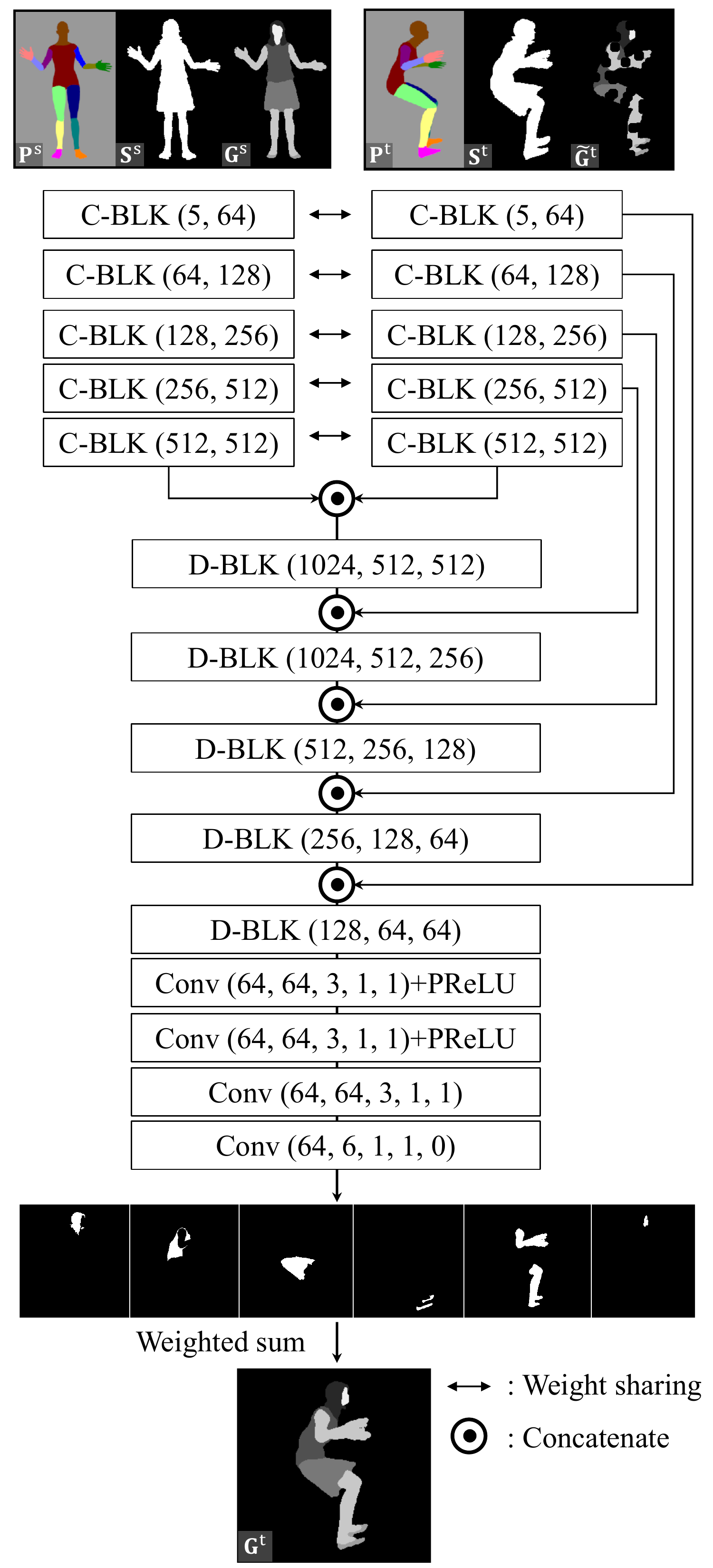}
\end{center}	
\vspace{-.5cm}
\caption{\small The details of our \emph{GarNet} implementation where C-BLK and D-BLK are described in Fig.~\ref{cov_block_network}. Conv and Deconv take as input parameters of (the number of input channels, the number of output channels, filter size, stride, the size of zero padding). We use 0.2 for the LeakyReLU (LReLU) coefficient.}
\label{garnet_network}
\end{figure}

We provide the implementation details of each modular function in our compositional pose transfer network. Fig.~\ref{silnet_network} describes the \emph{SilNet} architecture which takes as input source triplet of the pose map, garment labels, and silhouette, and target pose map, and predicts the silhouette mask in the target pose. Fig.~\ref{garnet_network} describes the architecture of our \emph{GarNet} that takes as input source triplet of the pose map, silhouette, and garment labels, and target triplet of the pose map, predicted silhouette, and pseudo garment labels, and predicts the complete garment labels. In Fig.~\ref{rendernet_network}, we show the details of our \emph{RenderNet} which takes as input source triplet of image, silhouette mask, and garment labels, target silhouette and garment labels, and target pseudo image and its mask, and generates the person image.

\section{More Results}\label{more_res}
\subsection{Additional Dataset Description}
We provide more details of the videos used for the evaluation. In order to evaluate our approach, we use eight sequences of the subjects in various clothing and motions from existing works~\cite{shimada2020physcap,yoon2020dynamic,liu2019liquid,alldieck2018video,habermann2020deepcap} and capture two more sequences which include a person with more complex clothing style and movement than others. \textit{RoM1} and \textit{RoM2}: Two men show their range of motion with various poses~\cite{shimada2020physcap}. \textit{Jumping}~\cite{yoon2020dynamic}: A woman in a black and white coat jump from one side to another. \textit{Kicking} and \textit{Onepiece}~\cite{habermann2020deepcap}: A man and woman take the motion of kicking and dancing where the woman is wearing a unique onepiece. \textit{Checker}~\cite{liu2019liquid}: A man in shirts with checkered pattern swings his hands. \textit{Rotation1} and \textit{Rotation2}~\cite{alldieck2018video}: Two A-posed men rotate their body. \textit{Maskman}: A man wearing a facial mask shows his various motion. \textit{Rainbow}: A woman in a sweater with rainbow pattern turns her body with dancing motion.
\subsection{User Study Results}
\begin{figure}
	\begin{center}
    \includegraphics[width=1\linewidth]{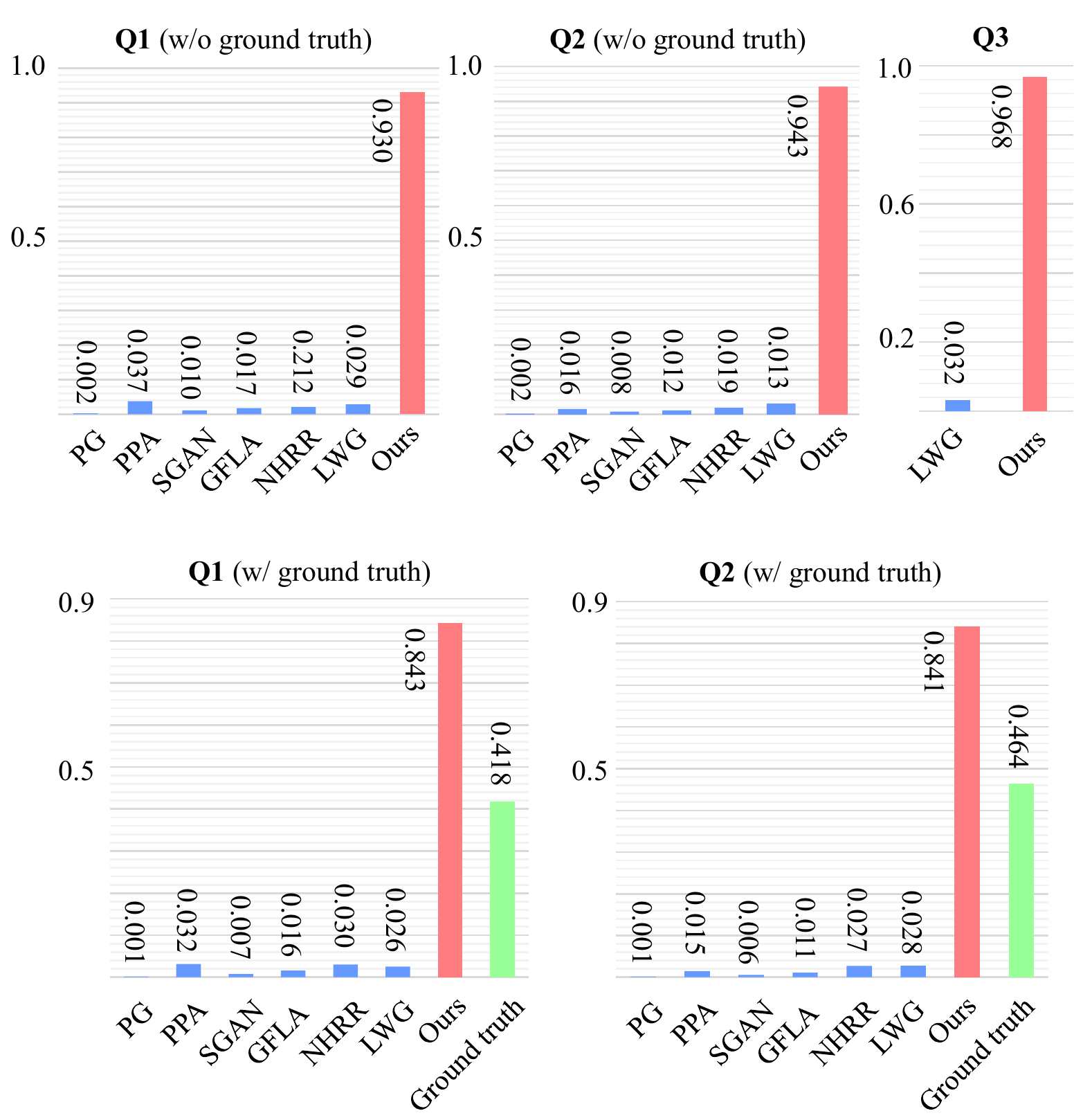}
	\end{center}	
	\vspace{-.5cm}
	\caption{\small The full results of the user study where \textit{x}-axis represents the number of votes for the associated method which is normalized by the number of participants and the number of occurrence in the questionnaires. Q1, Q2, and Q3 represent the question type. Our results were often ranked as more realistic than the real videos because they involve a significant boundary noise from the person segmentation error while our method produces the human animation with clean boundary.
	}
	\label{user_study}
\end{figure}

In our user study, three questions are asked: 
Q1: Which video looks most realistic including temporal coherence? Q2: Which video preserves the identity best including facial details, shape, and overall appearance? Q3: In which video, the background is preserved better across the frames (only for the case of scenes with background)? For each method, we measure the performance based on the number of entire votes divided by the number of participants and the number of occurrence in the questionnaires. 
The full results are shown in  Fig.~\ref{user_study}. 
%
%
The first question was answered in 84.3$\%$ and 93.0$\%$ of the cases in favour of our method with and without the ground truth sequence, respectively, and the second question 84.1$\%$ and 94.2$\%$. In the third question, the background is preserved better in our method than LWG in 96.8 $\%$ of the answers. The results show that our method outperforms other state of the art, and our animations are in many cases qualitatively comparable to real videos of the subjects. The choice between a real video and our animation did not fall easy because the ground-truth video often contains noisy boundary originated from the person segmentation error while the generated person images from our method shows the clear boundary.
\subsection{Additional Quantitative Results}
We include the quantitative results which do not appear in the main paper. In Table~\ref{table:eval}, the performance of the baseline models that are pretrained from the DeepFashion (DF) dataset by the authors is summarized in the first chunk (from 2 to 6 row), ablation study in the second chunk (from 7 to 16), and application to the multiview data in the third chunk (from 17 to 18).

\begin{table*}[t]
\centering
\small
\vspace{-1mm}
\begin{tabular}{|l|c|c|c|c|c|c|c|c|c|c|c|c}
\hline
 & \scriptsize{Maskman}  & \scriptsize{Rainbow}  & \scriptsize{RoM1}  & \scriptsize{RoM2} &  \scriptsize{Jumping} & \scriptsize{Kicking} & \scriptsize{Onepiece}  & \scriptsize{Checker} & \scriptsize{Rotation1}& \scriptsize{Rotation2} & \scriptsize{Average}   \\
\hline
\scriptsize{PG (DF)}&\scriptsize{2.01 / 4.24} &\scriptsize{2.14 / 4.41}  &\scriptsize{2.22 / 4.48}   &\scriptsize{1.81 / 4.31}    &  \scriptsize{2.33 / 4.32}&\scriptsize{2.15 / 4.49}  &  \scriptsize{2.43 4.66}&  \scriptsize{2.07 / 4.25}&\scriptsize{1.74 / 4.18}&\scriptsize{2.58 / 4.47}&\scriptsize{2.15 / 4.38}     \\
\hline

\scriptsize{SGAN (DF)}&\scriptsize{2.33 / 3.96}  &\scriptsize{2.39 / 4.22}   &\scriptsize{2.50 / 4.16}  &\scriptsize{2.12 / 4.22}  &\scriptsize{2.63 / 4.09}  &\scriptsize{2.49 / 4.29}  &\scriptsize{2.67 / 4.25}  &\scriptsize{2.34 / 3.99}  & \scriptsize{1.89 / 3.93} & \scriptsize{2.74 / 4.22} & \scriptsize{2.43 / 4.13} \\ 
\hline

\scriptsize{PPA (DF)}&\scriptsize{2.84 / 3.76}  &\scriptsize{2.70 / 3.80}   &\scriptsize{2.78 / 3.91}  &\scriptsize{2.65 / 3.97}  &\scriptsize{2.89 / 3.87}  &\scriptsize{2.88 / 3.94}  &\scriptsize{3.21 / 4.05}  &\scriptsize{2.26 / 3.76}  & \scriptsize{2.26 / 3.75} & \scriptsize{3.01 / 3.77} & \scriptsize{2.74 / 3.86} \\ 
\hline

\scriptsize{GFLA (DF)}&\scriptsize{1.96 / 3.86}  &\scriptsize{1.64 / 3.93} &\scriptsize{2.19 / 3.89}  &\scriptsize{1.50 / 3.99}  &  \scriptsize{2.01 / 3.85}  &\scriptsize{2.05 / 3.96}  &\scriptsize{2.23 / 3.94}  &\scriptsize{1.74 / 3.84}   &\scriptsize{1.60 / 3.88}&\scriptsize{1.92 / 3.89}&\scriptsize{1.88 / 3.90} \\ 
\hline

\scriptsize{NHHR (DF)}&\scriptsize{1.71 / 2.96}  &\scriptsize{1.89 / 3.06}  & \scriptsize{1.82 / 3.07} & \scriptsize{1.56 / 3.03} &   \scriptsize{2.06 / 3.03} & \scriptsize{1.68 / 3.11} & \scriptsize{2.16 / 3.16} & \scriptsize{1.48 / 2.94} & \scriptsize{1.80 / 3.02}  & \scriptsize{2.77 / 3.11}  & \scriptsize{1.89 / 3.05} \\ 


\hline
\hline

\scriptsize{R}& \scriptsize{1.64 / 2.31} &\scriptsize{1.48 / 2.43}  & \scriptsize{1.41 / 2.30} & \scriptsize{1.53 / 2.44} &   \scriptsize{2.00 / 2.54} & \scriptsize{1.16 / 2.18} & \scriptsize{1.36 / 2.34} & \scriptsize{1.41 / 2.31} &\scriptsize{1.22 / 2.31}&\scriptsize{1.62 /     2.33}&\scriptsize{1.48 / 2.35} \\ 
\hline
\scriptsize{GR}& \scriptsize{1.64 / 2.30} & \scriptsize{1.45 / 2.42}& \scriptsize{1.51 / 2.30} & \scriptsize{1.44 / 2.42} &   \scriptsize{1.91 / 2.53} & \scriptsize{1.40 / 2.24} & \scriptsize{1.24 / 2.35} & \scriptsize{1.39 / 2.29} &\scriptsize{1.21 / 2.30}&\scriptsize{1.60 / 2.32}&\scriptsize{1.47 / 2.35}  \\ 
\hline
\scriptsize{SR}& \scriptsize{1.57 / 2.26}  &\scriptsize{1.30 / 2.42} &\scriptsize{1.31 / 2.24}  &\scriptsize{1.41 / 2.37}   &\scriptsize{1.89 / 2.54}  &\scriptsize{1.17 / 2.20}  &\scriptsize{1.24 / 2.33}    &\scriptsize{1.11 / 2.24} &\scriptsize{1.05 / 2.23} &\scriptsize{1.25 / 2.22} &\scriptsize{1.33 / 2.31}\\ 
\hline
\scriptsize{SGR-$\mathbf{S}^{\rm s}$}& \scriptsize{1.58 / 2.29}  &\scriptsize{1.33 / 2.41} &\scriptsize{1.26 / 2.26}  &\scriptsize{1.43 / 2.39}   &\scriptsize{1.99 / 2.54}  &\scriptsize{1.18 / 2.23}  &\scriptsize{1.29 / 2.35}  &\scriptsize{1.10 / 2.36}  &\scriptsize{1.05 / 2.23} &\scriptsize{1.24 / 2.20} &\scriptsize{1.35 / 2.32} \\ 

\hline
\scriptsize{SGR-$\mathbf{G}^{\rm s}$}&  \scriptsize{1.66 / 2.30} & \scriptsize{1.38 / 2.39}& \scriptsize{1.31 / 2.32} & \scriptsize{1.48 / 2.35}  &\scriptsize{1.89 / 2.51}  &\scriptsize{1.18 / 2.23}  &\scriptsize{1.31 / 2.40}  &\scriptsize{1.31 / 2.31}  &\scriptsize{1.19 / 2.28}&\scriptsize{1.42 / 2.30}&\scriptsize{1.41 / 2.34}\\ 
\hline
\scriptsize{SGR-$\bar{\mathbf{I}}^{\rm t}$}&  \scriptsize{1.79 / 2.28} & \scriptsize{1.97 / 2.49}& \scriptsize{1.55 / 2.30} & \scriptsize{1.52 / 2.38}  &\scriptsize{2.13 / 2.50}  &\scriptsize{1.31 / 2.23}  &\scriptsize{1.79 / 2.39}  &\scriptsize{1.49 / 2.31}  &\scriptsize{1.15 / 2.22}&\scriptsize{1.50 / 2.21}&\scriptsize{1.62 / 2.33}\\

\hline
\scriptsize{SGR-$\mathbf{z}^{\rm s}$}& \scriptsize{1.57 / 2.27}  &\scriptsize{1.31 / 2.40} &\scriptsize{1.25 / 2.26}  &\scriptsize{1.42 / 2.38}   &\scriptsize{1.90 / 2.52}  &\scriptsize{1.15 / 2.19}  &\scriptsize{1.29 / 2.31}  &\scriptsize{1.11 / 2.22}  &\scriptsize{1.05 / 2.19} &\scriptsize{1.24 / 2.23} &\scriptsize{1.32 / 2.30} \\ 

\hline
\scriptsize{SGR-$L_{\rm KL}$}& \scriptsize{1.54 / 2.27}  &\scriptsize{1.25 / 2.38} &\scriptsize{1.27 / 2.25}  &\scriptsize{1.40 / 2.38}   &\scriptsize{1.88 / 2.55}  &\scriptsize{1.13 / 2.19}  &\scriptsize{1.25 / 2.32}  &\scriptsize{1.09 / 2.24}  &\scriptsize{1.04 / 2.20} &\scriptsize{1.15 / 2.19} &\scriptsize{1.30 / 2.30} \\ 

\hline
\scriptsize{SGR-$\mathbf{A}$}& \scriptsize{1.59 / 2.28}  &\scriptsize{1.28 / 2.40} &\scriptsize{1.31 / 2.26}  &\scriptsize{1.40 / 2.38}   &\scriptsize{1.86 / 2.51}  &\scriptsize{1.23 / 2.21}  &\scriptsize{1.32 / 2.33}  &\scriptsize{1.14 / 2.25}  &\scriptsize{1.15 / 2.23} &\scriptsize{1.28 / 2.20} &\scriptsize{1.36 / 2.31} \\

\hline
\scriptsize{SGR (\textbf{full})}&  \scriptsize{\textbf{1.54} / \textbf{2.27}} & \scriptsize{\textbf{1.24} / \textbf{2.38}}& \scriptsize{\textbf{1.25} / \textbf{2.24}} & \scriptsize{\textbf{1.38} / \textbf{2.36}}  &\scriptsize{\textbf{1.87} / \textbf{2.53}}  &\scriptsize{\textbf{1.08} / \textbf{2.19}}  &\scriptsize{\textbf{1.23} / \textbf{2.32}}  &\scriptsize{\textbf{1.09} / \textbf{2.24}}  &\scriptsize{\textbf{1.00} / \textbf{2.19}}&\scriptsize{\textbf{1.12} / \textbf{2.16}}&\scriptsize{\textbf{1.28} / \textbf{2.29}}\\ 

\hline
\hline

\scriptsize{SGR+2view}&  \scriptsize{1.50 / 2.25} & \scriptsize{1.22 / 2.38}& \scriptsize{1.21 / 2.23} & \scriptsize{1.33 / 2.36}  &\scriptsize{1.80 / 2.51}  &\scriptsize{1.15 / 2.17}  &\scriptsize{1.20 / 2.31}  &\scriptsize{1.07 / 2.23}  &\scriptsize{0.97 / 2.16}&\scriptsize{1.06 / 2.14}&\scriptsize{1.25 / 2.28}\\ 
\hline
\scriptsize{SGR+4view}&  \scriptsize{1.49 / 2.25} & \scriptsize{1.21 / 2.38}& \scriptsize{1.21 / 2.23} & \scriptsize{1.33 / 2.35}  &\scriptsize{1.80 / 2.51}  &\scriptsize{1.12 / 2.17}  &\scriptsize{1.20 / 2.31}  &\scriptsize{1.07 / 2.23}  &\scriptsize{0.98 / 2.16}&\scriptsize{1.07 / 2.14}&\scriptsize{1.24 / 2.27}\\ 

\hline
\end{tabular}
\vskip-7pt
\caption{\small Quantitative results with LPIPS (left, scale: $10^{\shortminus1}$) and CS where the lower is the better. We denote the full model used for the comparison with other baseline methods as SGR (\textbf{full}).}
\label{table:eval}
\vskip-6pt
\end{table*}




\begin{figure}
	\begin{center}
    \includegraphics[width=1\linewidth]{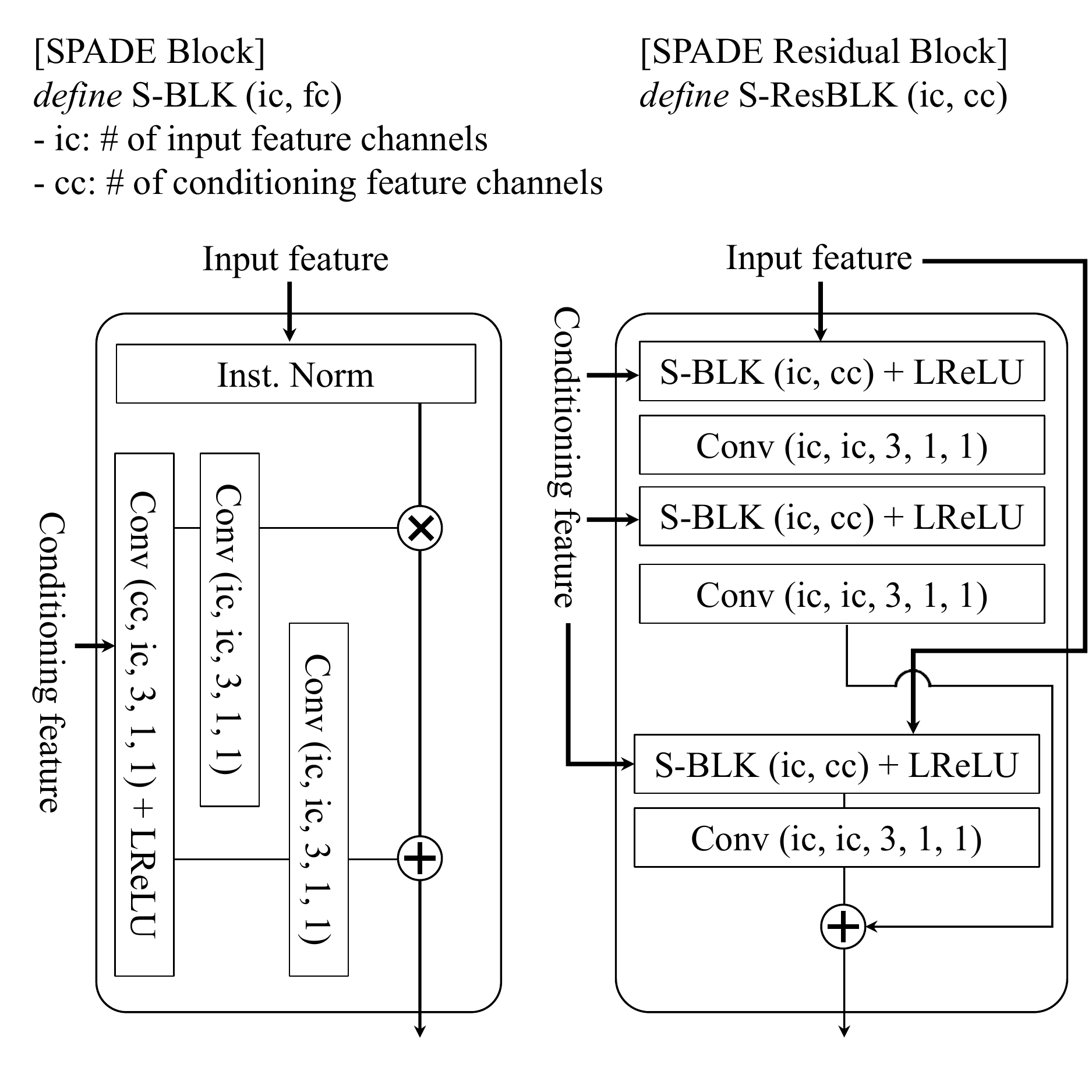}
	\end{center}	
	\vspace{-.5cm}
	\caption{\small The description of SPADE and SPADE Residual blocks similar to~\cite{park2019semantic}. Conv take as input parameters of (the number of input channels, the number of output channels, filter size, stride, the size of zero padding). We use 0.2 for the LeakyReLU (LReLU) coefficient.
	}
	\label{spade_block}
\end{figure}

\begin{figure}
	\begin{center}
    \includegraphics[width=1\linewidth]{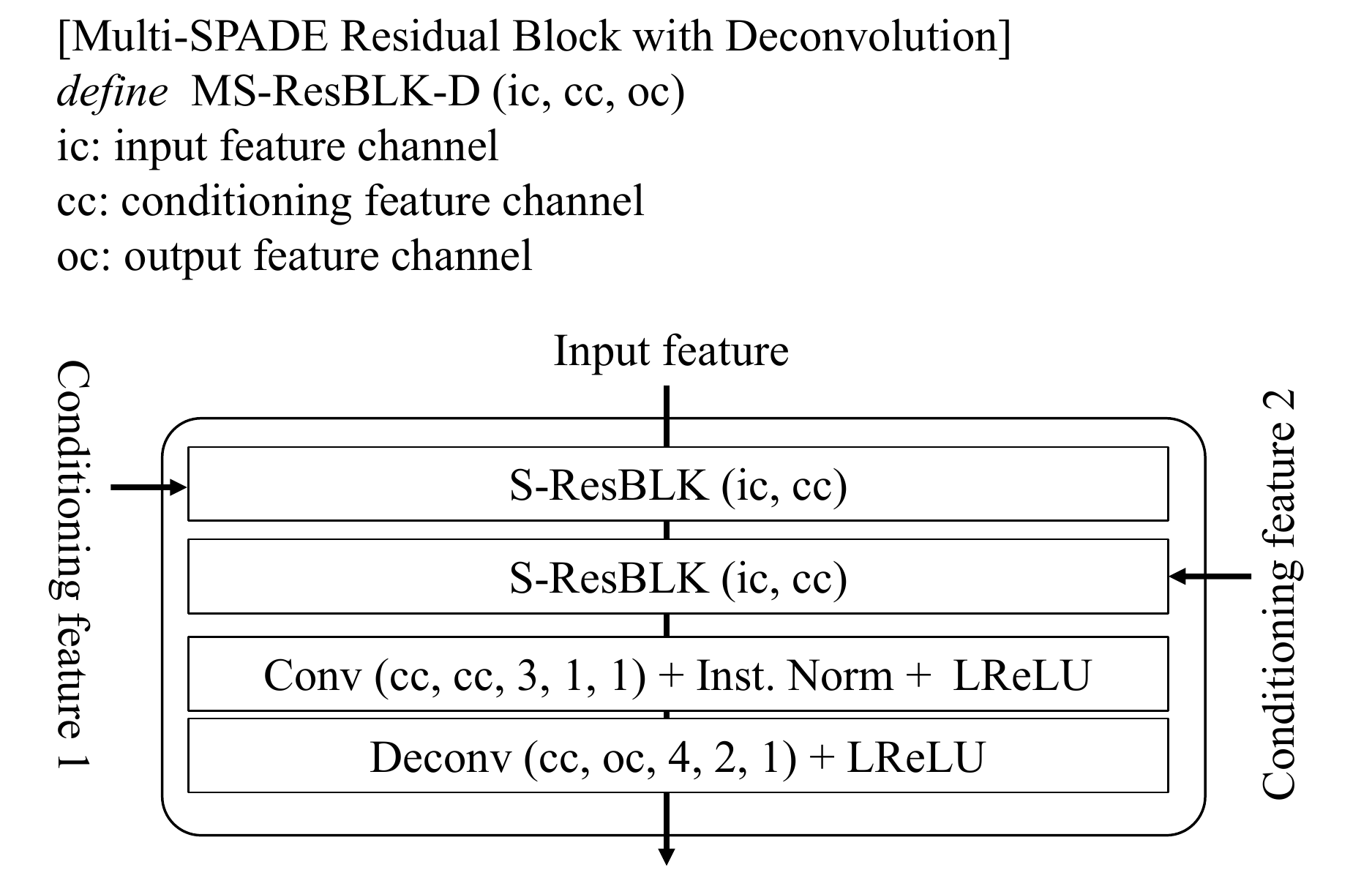}
	\end{center}	
	\vspace{-.5cm}
	\caption{\small The description of Multi-Spade blocks similar to~\cite{mallya2020world} where the details of S-ResBLK is described in Fig.~\ref{spade_block}.  Conv and Deconv take as input parameters of (the number of input channels, the number of output channels, filter size, stride, the size of zero padding). We use 0.2 for the LeakyReLU (LReLU) coefficient. 
	}
	\label{mspade_block}
\end{figure}

\begin{figure}
	\begin{center}
    \includegraphics[width=1\linewidth]{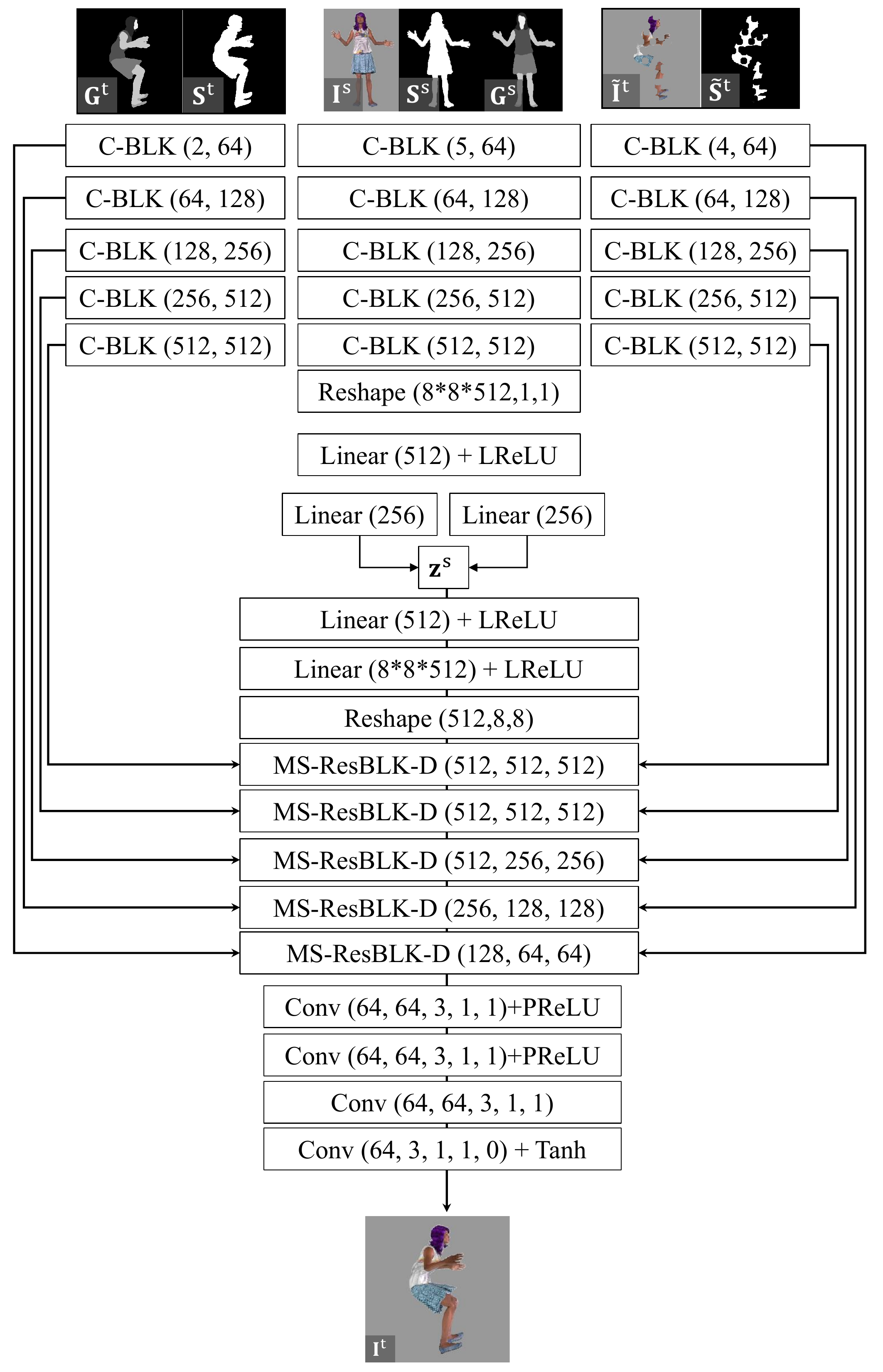}
	\end{center}	
	\vspace{-.5cm}
	\caption{\small The details of our \emph{RenderNet} where C-BLK and D-BLK are described in Fig~\ref{cov_block_network}, and MS-ResBLK-D is in Fig.~\ref{mspade_block}.  Conv takes as input parameters of (the number of input channels, the number of output channels, filter size, stride, the size of zero padding). We use 0.2 for the LeakyReLU (LReLU) coefficient.
	}
	\label{rendernet_network}
\end{figure}

\end{document}